\documentclass[letterpaper]{article} 
\usepackage{aaai2027}

\usepackage[hyphens]{url}  
\usepackage{graphicx} 
\usepackage{natbib}  
\usepackage{caption} 
\usepackage{amsmath}
\usepackage{amssymb}
\usepackage{amsfonts}
\usepackage{amsthm}
\usepackage{mathtools}
\usepackage{placeins}
\usepackage{booktabs}
\usepackage{multirow}
\usepackage{array}
\usepackage{tablefootnote}
\usepackage{threeparttable}
\usepackage{longtable}

\usepackage{algorithm}
\usepackage{algorithmic}

\usepackage{tabularx}
\usepackage{newfloat}
\usepackage{listings}
\DeclareCaptionStyle{ruled}{labelfont=normalfont,labelsep=colon,strut=off} 
\floatstyle{ruled}
\newfloat{listing}{tb}{lst}{}
\floatname{listing}{Listing}

\title{Teaching AI, Robotics, \& Community: \\A Hubs-Based K--12 Education Framework for Reaching Rural Schools}

\author{
Maxwell J. Jacobson,
Gustavo Rodriguez-Rivera,
Petros Drineas,
and Yexiang Xue
}

\affiliations{
Purdue University\\
jacobs57@purdue.edu,
grr@purdue.edu,
pdrineas@purdue.edu,
yexiang@purdue.edu

}

\begin{document}

\maketitle

\begin{abstract}
K--12 robotics and AI education remains difficult to scale, especially in rural regions lacking sustained technical mentorship. Programs like FIRST provide competition pathways and instructional opportunities, but they do not eliminate the need for local programming and robotics expertise. We introduce AI, Robotics, \& Community (ARC), a hubs-based framework where colleges train undergraduate mentors and host workshops for nearby K--12 teams. Mature school programs can become secondary hubs that support additional schools, creating a self-reinforcing education loop where mentorship reach propagates geographically and can even grow super-linearly. We first evaluate ARC through a trial deployment at one university. The trial created three rural robotics teams. On five-point Likert surveys, mean increases in K--12 programming knowledge, resource access, and practice opportunities were 2.00, 2.25, and 1.25 points. Likewise, undergraduate confidence teaching technical concepts, adapting explanations, managing groups, and finding mentoring enjoyable and meaningful increased by 1.29, 1.14, 1.00, and 1.14 points. Additionally, we create a spatial Markov model of ARC's growth and simulate it using the state of Indiana as a testbed. Under moderate conditions, we find that ARC reaches 74\% of Indiana's 1,925 public K--12 schools and produces 992 robotics programs after 40 years, compared with 161 projected under natural growth alone. Together, these results show ARC can create and support rural robotics programs, train undergraduate AI and robotics mentors, and potentially scale mentorship across a region.
\end{abstract}

\begin{figure}[tb]
    \centering
    \includegraphics[width=1\linewidth]{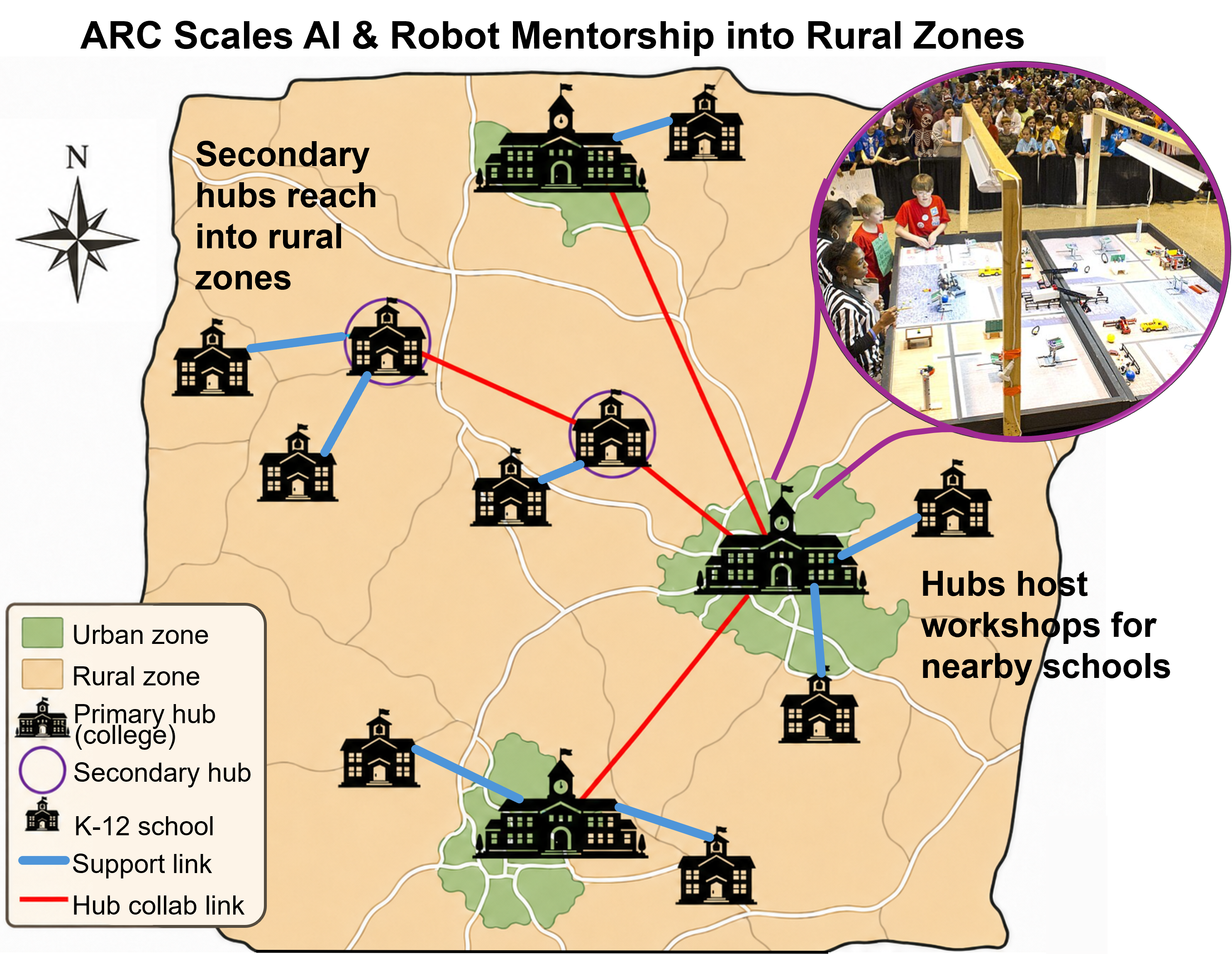}
    \caption{Overview of the ARC hubs-based education framework. Primary hubs at colleges train undergraduate mentors in robotics and AI and host workshops for nearby K--12 robotics teams. As programs mature, some become secondary hubs that extend robotics and AI mentorship into rural areas and support additional nearby schools, allowing the network to propagate beyond the reach of universities.}
    \label{fig:arc_main}
    \vspace{-10pt}
\end{figure}


\begin{figure*}[tb]
    \centering
    \includegraphics[width=0.9\linewidth]{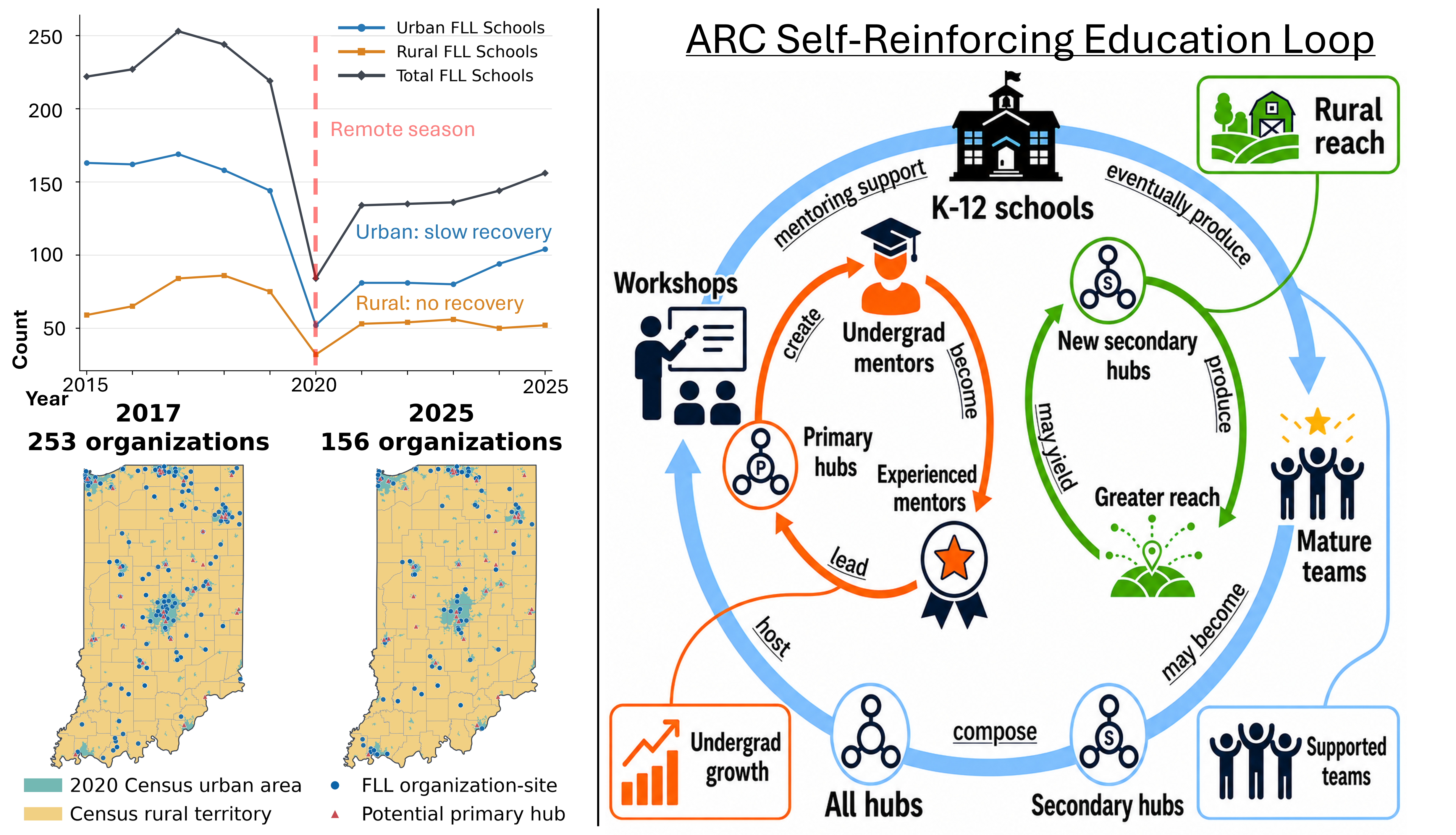}
    \caption{\textbf{Left:} FIRST LEGO League participation in Indiana before and after the 2020 remote season participation crash. The left panel shows urban, rural, and total FLL organizations from 2015--2025. The maps show the geographic distribution of participating organizations in 2017 (the high point) and 2025 relative to Census urban areas and major STEM universities. Urban participation has gradually recovered, while rural participation has not seen a recovery.
    \textbf{Right:} ARC creates a self-reinforcing education loop. The primary loop (blue) shows how hubs host workshops that support K–12 schools, whose teams can mature into secondary hubs that expand the overall hub network and enable additional workshops. Zooming in, the undergraduate growth loop (red) shows how primary hubs create undergraduate mentors, who can become experienced mentors and later lead or support primary hubs, allowing those hubs to train additional undergraduates. Finally, the rural growth loop (green) shows how new secondary hubs expand geographic reach, and how that greater reach may yield additional secondary hubs. These loops integrate at multiple points, reinforcing themselves and each other to produce three desirable outcomes: more team support, more undergraduate growth, and greater reach into rural areas.}
    \label{fig:arc_promblem_and_loops}
    \vspace{-10pt}
\end{figure*}

\section{Introduction}

Scaling AI and robotics literacy in K--12 education is difficult, especially in under-served rural regions. Robotics and AI are increasingly important to future economies, innovation, and civic life, but access to meaningful robotics and AI learning is uneven for schools that lack sustained technical support. Indiana provides an example of this access gap, where rural participation in the FIRST LEGO League (FLL) robotics competition has recovered much more slowly than urban participation following the 2020 remote season. Without early exposure, students may be less prepared to understand how AI and robots work and may become passive consumers of technology rather than active creators or critical evaluators.

Existing robotics programs provide a strong foundation for K--12 robotics education. In particular, the FIRST ecosystem offers a progression 
of team based competitions,
giving students opportunities to build, program, and compete with robots at increasing levels of complexity. Schools, teachers, and community volunteers already support these programs through local clubs and competitions. Related work has also developed curricula for teaching AI concepts and ethics in middle and high school \cite{williams2021robot,alvarez2022socially,krakowski2022authentic}.

The missing piece is not the absence of robotics programs or educational material, but the difficulty of scaling sustained technical mentorship. Many K--12 teams depend on a small number of volunteers, and geographically dispersed rural schools may have limited access to mentors with programming, robotics, or AI expertise. A school may have interested students and a willing teacher while still lacking the technical knowledge needed to start and sustain a program. This mentorship gap limits the formation, persistence, and growth of K--12 robotics teams. Students in under-supported regions therefore have fewer opportunities to develop programming, engineering, and AI literacy even when interest exists. At the same time, undergraduate computer science students often have few structured opportunities to deepen their own technical understanding through teaching and to apply their knowledge in real educational settings.

We propose the AI, Robotics, \& Community education framework (ARC) for scaling K--12 robotics and AI mentorship. ARC organizes hands-on \textit{workshop events} around regional hubs, allowing multiple teams in an area to receive direct technical mentorship. Each team works with a small number of mentors on the programming and robotics skills needed to make progress in the current competition. Workshop content is adapted to age and competition level, from programming and robot design for younger teams to autonomy, computer vision, and AI for advanced teams.

\textit{Primary hubs} are colleges or other STEM institutions that run the ARC course, train undergraduate computer science students as robotics and AI mentors, and host workshops for nearby K--12 teams. An experienced mentor prepares the undergraduates in technical skills as well as the educational and interpersonal skills needed to teach younger students. The undergraduates then lead the workshops under supervision. The course also exposes them to local research in AI and robotics and its effects on their communities. By helping students see technical mentorship as a meaningful role within their local community, ARC aims to encourage continued mentoring and volunteering after the course ends.

As K--12 teams become mature and self-sustaining, some can become secondary hubs that host workshops for nearby schools with limited assistance from the primary-hub network. Experienced K--12 students then become peer mentors for other teams. Secondary hubs allow technical support to extend beyond universities and farther into rural regions. More importantly, secondary hubs can help nearby teams mature into additional secondary hubs. Consequently, ARC creates a \textit{self-reinforcing education loop}: primary hubs train undergraduates, undergraduates teach K--12 students at workshops, K--12 teams become self-sustaining, and mature teams can then host their own workshops. Because each new secondary hub can help create further hubs, ARC has the potential for multiplicative and super-linear growth in mentorship capacity and geographic reach.

ARC addresses the mentorship gap by allowing technical support to spread geographically rather than remain concentrated around universities and urban zones. Primary hubs support nearby schools, while mature K--12 programs can become secondary hubs capable of supporting schools farther into rural regions. Schools therefore gain access to technical knowledge while also developing the capacity to eventually provide that knowledge to others.

We evaluate the growth potential of ARC using a spatially-explicit Markov model that represents both school robotics-program development and each school's relationship to the ARC network. In an Indiana simulation containing 1,925 public K--12 schools, a moderate set of ARC growth assumptions reaches approximately 74\% of schools and produces 992 schools with robotics programs after 40 years, compared with only 161 programs under the ARC-off natural-growth baseline. Under the most optimistic assumptions considered, ARC reaches approximately 95\% of the modeled state and produces approximately 1,638 school programs, or 85\% of Indiana public K--12 schools, with clear super-linear growth as secondary hubs multiply. Even under the least optimistic case considered, the model predicted 338 programs, more than twice the predicted number without our program.

We also conduct an initial deployment of ARC through one primary hub supporting three new FLL teams. In an undergraduate mentors survey measuring several axes of growth in a 5-point Likert score (strongly disagree to strongly agree), confidence teaching technical concepts increased from $3.00 \pm 1.15$ to $4.29 \pm 0.49$, confidence adapting explanations increased from $3.29 \pm 1.11$ to $4.43 \pm 0.53$, and connection to the local community increased from $1.86 \pm 1.07$ to $4.00 \pm 0.82$. Parents reported similarly large changes for participating students, including increases in basic robot programming knowledge from $2.00 \pm 1.41$ to $4.00 \pm 1.41$ and access to needed programming resources from $2.00 \pm 0.82$ to $4.25 \pm 0.50$. They also rated their children's enjoyment of working with the undergraduate mentors at $4.75 \pm 0.50$ and mentor support at $4.50 \pm 0.58$ out of five. Together, the simulation and trial deployment provide evidence that ARC can both produce effective local mentorship and, through repeated hub formation, potentially scale that mentorship across an entire region.

\section{The Rural Robotics Access Gap}


Programs like FIRST provide K--12 students with opportunities to learn programming, robotics, AI, and related skills through team-based competitions. FIRST provides several program levels -- FIRST LEGO League (FLL) introduces younger students to robotics and programming, FIRST Tech Challenge (FTC) provides a more advanced robotics competition for middle- and high-school students, and FIRST Robotics Competition (FRC) provides a larger-scale competition for high-school students. Together, these form a path for students to learn increasingly advanced skills.

However, access to these programs is geographically uneven. The upper left panel of Figure~\ref{fig:arc_promblem_and_loops} shows this pattern for FLL participation in Indiana. Participation fell sharply during the 2020 remote season in both urban and rural areas. Since then, urban participation has gradually recovered, with both new and returning schools participating. Rural participation has not shown the same recovery. In the 2025--2026 season, rural FLL participation remained close to its post-2020 level and well below its 2017 peak. The lower left panels of Figure~\ref{fig:arc_promblem_and_loops} show the corresponding geographic pattern, with much of the recovered participation concentrated in and around urban areas and major STEM campuses.

One important mechanism behind this gap is access to mentor resources. By mentor resources, we mean people with enough robotics, AI, and programming knowledge to help teams start, develop technical skills, and continue participating over time. Mentorship has already been identified as an important constraint on school robotics programs. Teacher knowledge can affect the adoption and continued use of educational robotics \cite{stokes2023robotics}, while robotics preparation may be particularly limited among rural educators \cite{meador2025rural}. This creates a difficult problem for rural schools. Even when students and teachers are interested in robotics, they may have limited access to the technical knowledge needed to sustain a program.

\section{A Hubs-Based K--12 Education Framework}

The AI, Robotics, \& Community (ARC) framework is a hubs-based approach for expanding access to robotics and AI mentorship across a region. This is accomplished via local workshop events run by \textit{hub} locations. \textit{Primary hubs} are colleges or other STEM institutions that train undergraduate mentors and host workshops for nearby K--12 robotics teams. As supported teams gain experience and become self-sustaining, some can become \textit{secondary hubs} that host workshops for other nearby schools. ARC therefore provides a way for technical mentorship to spread geographically rather than remain concentrated around universities. Because secondary hubs can help create additional secondary hubs, this structure also has the potential for super-linear growth in mentorship resources and geographic reach, allowing rural schools to be efficiently serviced.

\begin{figure}[tb]
    \centering
    \includegraphics[width=1\linewidth]{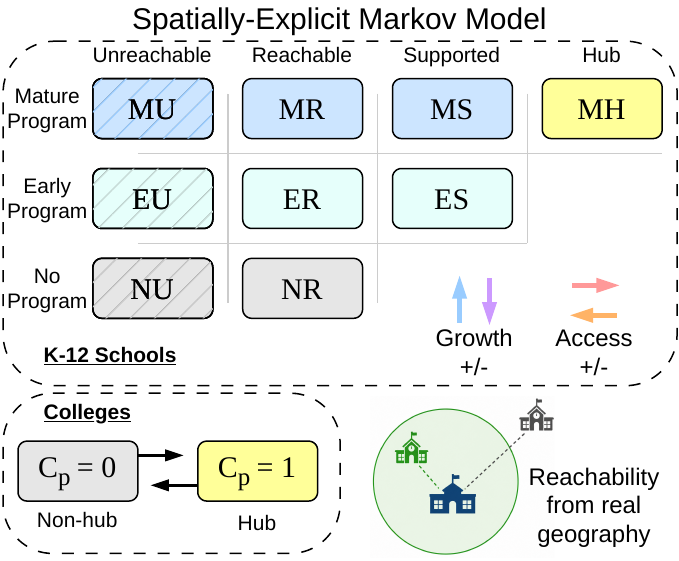}
    \caption{Simplified transition diagram for the spatially-explicit Markov model. The two axes of the K--12 school grid represent underlying state variables: program maturity $\in \{\text{no program}, \text{early program}, \text{mature program}\}$, and ARC level $\in \{\text{unreached}, \text{reached}, \text{supported}, \text{secondary hub}\}$. Each transition is a probability of changing states by year end. Colleges are modeled separately as hubs or non-hubs. Reachability depends on geographic distance to a hub -- the blue hub can support the green school, but not the gray one.
    Ablations can be tested by disabling model transitions (setting their probability to 0). For example, isolating the unreachable column (hatched) simulates expected growth without ARC.
    }
    \label{fig:transition_model}
    \vspace{-10pt}
\end{figure}

\subsection{Educational Workshops}

Workshops are the basic unit through which ARC provides mentorship to K--12 teams. They are hands-on and mentor-led, with several teams attending at the same time and each team working directly with a small group of mentors. Game arenas are provided so that students can work on their own robots and the tasks of the current year's competition. Rather than teaching programming or robotics in the abstract, mentors focus on the skills that students need to make progress in the game. For example, a mentor may focus on teaching the concepts needed to make a robot navigate to a ball, and push it stably towards a goal. Skills like motor control, use of sensors, planning, and logic may all follow from this. Students plan robot behavior, write and debug programs, test their solutions on the game arena, and revise them with direct help from their mentors. Workshop content is adapted to the age, experience, and competition level of each team. For FLL teams, this primarily involves foundational robotics and programming skills. FTC and FRC teams can require more advanced robotics, autonomy, computer vision, and AI depending on the needs of the game and the team. 

The workshop format makes mentorship and resource sharing more practical across geographically widespread regions. Instead of requiring experienced mentors to travel repeatedly to many individual schools, several teams can travel to a nearby hub and receive focused support simultaneously. A limited number of mentors can support more teams while still giving each team direct, hands-on assistance.

\subsection{The ARC Course at Primary Hubs}

Primary hubs are colleges or other institutions that run the ARC course and use trained undergraduate students to provide workshops. The course gives the hub a regular source of mentors, as well as a specific time and place for workshops to occur. An experienced mentor trains the undergraduate students and oversees their work with the K--12 teams. The undergraduates then lead the workshops, working directly with individual teams to teach programming, robotics, debugging, and other skills needed for the competition. Mentor preparation therefore includes both technical knowledge and the interpersonal and educational skills needed to communicate technical ideas to students at different K--12 levels.
The course begins with focused preparation in robotics and programming, robotics education, youth protection, and methods for teaching programming. After this preparation, undergraduates lead supervised K--12 workshops, with short continuing seminars connecting their mentoring to AI, robotics, and community applications. Full course details and reproducibility recommendations are provided in Appendix~\ref{sec:course_reproducibility}.


\subsection{Mature Schools as Secondary Hubs}

Secondary hubs extend the same workshop model beyond the primary hubs. A secondary hub is a mature K--12 school with an active, self-sustaining robotics program and enough internal technical knowledge to help nearby teams. This builds on the existing FIRST culture of experienced teams helping less experienced teams. At a secondary-hub workshop, experienced K--12 students become the main peer mentors for participating teams. A teacher or a small number of experienced undergraduate mentors can provide oversight and additional technical support when needed.

Secondary hubs are especially important for extending ARC into rural areas. A school may be too far from a university for repeated participation at a primary hub, but close enough to another K--12 school to attend workshops there. As teams around a secondary hub receive mentorship and mature, some may eventually become secondary hubs themselves. Those new hubs can then support another set of nearby schools in an expanded region of reachability. ARC can then expand through repeated local growth as well as the creation of additional university programs. The ability of secondary hubs to create more secondary hubs is the mechanism that gives the framework its potential for super-linear growth.

\section{Spatially-Explicit Markov Model of Growth}

\begin{figure*}[tb]
    \centering
    \includegraphics[width=0.95\linewidth]{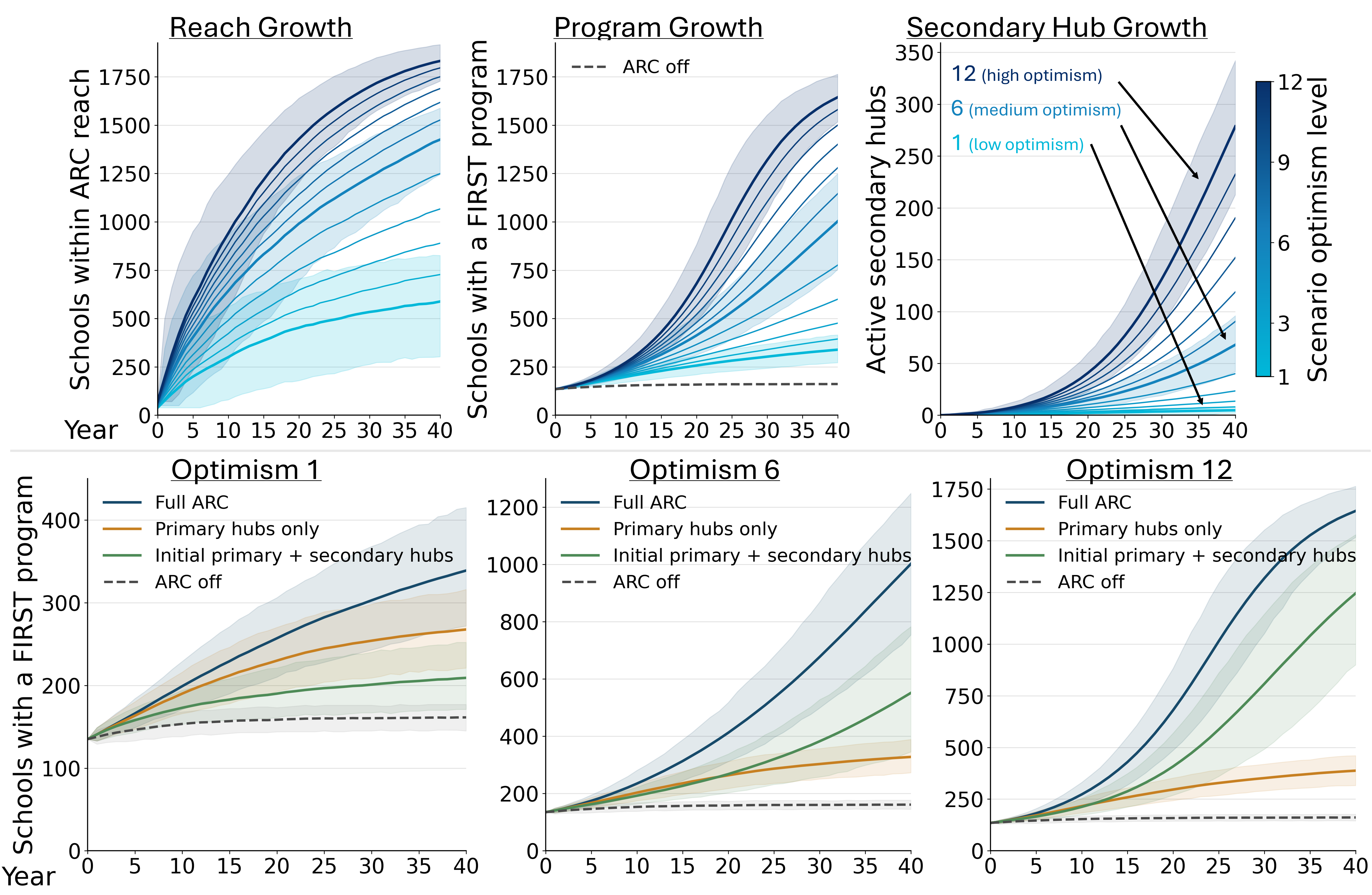}
    \caption{Simulation results for 12 \textit{optimism level} ARC growth scenarios. Depending on these parameter sets, the model shows that \textbf{superlinear program growth is possible.}
    \textbf{Top:} the number of schools reached (left), programs existing (middle), and secondary hubs existing (right) over 40 year simulations. Shaded regions show 10th--90th percentile ranges over 1000 runs. The more optimistic scenarios show pronounced self-reinforcing takeoffs in both secondary hubs and school programs. The less optimistic ones do not, but still produce much more rapid growth than projections without ARC. 
    \textbf{Bottom:} model ablation at scenarios 1, 6, and 12. Primary hubs alone increase growth but remain baseline-shaped. Secondary hubs show accelerating growth, though slowly from a single starting hub. Full ARC uses both, enabling parallel expansion and the fastest growth.}
    \label{fig:arc_growth_charts}
    \vspace{-10pt}
\end{figure*}

To examine how ARC could grow across a region, we model its development as a spatially-explicit Markov process with annual transitions. We consider a mixed-rural region containing $n$ K--12 schools and $m$ colleges or other institutions that could operate as primary hubs. The goal of ARC is to expand access to robotics and programming knowledge across the K--12 schools in this region. The model represents both the natural development of school robotics programs and the additional effects of ARC support and hub growth.

Each potential college primary hub $p$ has a state $C_p(t)\in\{0,1\}$ indicating whether it is inactive or currently operating as an ARC primary hub. At each annual transition, an inactive college may become a primary hub, while an active primary hub may cease operating. Active primary hubs extend ARC's geographic reach in large steps and contribute capacity for launching and supporting school programs. The model does not attempt to represent the internal process by which a college decides to begin or end participation. These changes are instead represented by transition probabilities that may vary across colleges and time.

K--12 schools have two state variables, shown in Figure~\ref{fig:transition_model}. The FIRST program state $F_i(t)\in\{N,E,M\}$ describes whether school $i$ has no program, an Early program, or a Mature program. The ARC state $L_i(t)\in\{U,R,S,H\}$ describes whether the school is Unreachable by the current hub network, Reachable, directly Supported, or itself a secondary Hub. The school program is the atomic unit of the model rather than an individual team. Not every combination of these variables is possible. A Supported school must have an active program, and a secondary Hub must have a Mature program, leaving nine legal school states. 
Annual transitions capture the ways in which both school programs and the ARC network can change. Programs can naturally start, mature, close, or backslide from Mature to Early. ARC can help launch a program at a reachable school, directly support an existing program, and increase the opportunity for a Mature school to become a secondary hub. Hubs themselves can also stop operating. Geography enters through a fixed relation specifying which schools each potential primary or secondary hub can serve. The active hub network therefore determines which schools are reachable, and new secondary hubs can expand that reach. Reachability alone does not alter a school's natural program-development probabilities. Direct ARC support is what allows different maturation and persistence behavior. The complete transition model is given in Appendix~\ref{sec:model_full}. A mean-field approximation that replaces exact geography and individual institutions with regional averages is provided in Appendix~\ref{sec:model_mf} -- this is more analytically tractable but loses spatial clustering, overlapping coverage, and local capacity constraints.

For our analysis, we instantiate the full spatial model for Indiana, with 1,925 public K--12 schools and 32 colleges with computer science programs treated as potential primary hubs. Initial program states are based on observed FIRST participation, and ARC-specific quantities not yet directly measurable are varied across 12 \textit{optimism levels} with monotonically more favorable assumptions. Each level is simulated for 40 years over 1,000 Monte Carlo runs. We also form three ablations by setting the incoming transition probabilities for selected parts of ARC to zero. \textit{ARC off} disables ARC growth entirely, leaving only natural program dynamics. \textit{Primary hubs only} disables secondary-hub formation while allowing additional primary hubs. \textit{Initial primary + secondary hubs} starts from the initial primary hub and allows secondary hubs to form, but prevents additional primary hubs from activating. These isolate the contributions of the two hub-growth mechanisms from \textit{Full ARC}. Complete parameterization and extended results are provided in Appendices~\ref{sec:model_analysis} and \ref{sec:extended_sim_results}.

The full spatial simulations show substantial growth across the modeled range (Figure~\ref{fig:arc_growth_charts}). After 40 years, \textit{ARC off} produces a mean of 161 programs, while even optimism level 1 of \textit{Full ARC} produces 338, more than twice as many. At level 6, the model produces 992 programs and reaches 1,415 schools. At level 12, 1,638 schools have programs and 1,829 are within ARC reach, approximately 85\% and 95\% of Indiana public K--12 schools, respectively. Thus the strongest modeled conditions produce more than ten times as many programs as natural growth alone. Level 6 reaches 61\% of rural schools and yields 341 rural programs, compared with 60 under ARC off, and 118 with primary hubs alone.

The trajectory shapes support ARC's proposed self-reinforcing growth mechanism. Lower optimism levels increase program growth without a strong takeoff, while higher levels show accelerating growth as secondary hubs multiply. This is clearest in the secondary-hub and program trajectories in Figure~\ref{fig:arc_growth_charts}, where the more favorable scenarios develop pronounced superlinear growth before eventually slowing as the finite school population begins to saturate.

The ablations show that this growth is not produced by either hub mechanism alone. \textit{Primary hubs only} expands access across multiple regions but remains comparatively baseline-shaped, while \textit{Initial primary + secondary hubs} can generate accelerating growth but must propagate outward from a single starting region. At optimism level 6, these models produce 323 and 551 programs after 40 years, respectively, compared with 992 under \textit{Full ARC}. Primary hubs therefore provide geographically distributed starting points while secondary hubs recursively extend them, with the combined ARC design producing substantially stronger growth than either mechanism in isolation.

We additionally tested the ARC-specific assumptions using one-at-a-time sweeps and Morris sensitivity screening, with full methods and results in Appendix~\ref{sec:parameter_sensitivity}. Both tests show that early program growth is driven primarily by primary-hub activation and secondary-hub formation from existing Mature programs, while later growth shifts toward Supported schools becoming new secondary hubs, together with direct support, reach, and capacity. At year 40, varying Supported secondary-hub formation, support capacity, and reach radius across their optimism-level ranges changed projected program counts by approximately 430, 338, and 302 schools, with Morris likewise ranking Supported secondary-hub formation as the strongest long-term driver.

\begin{figure}[t]
    \centering
    \includegraphics[width=0.85\linewidth]{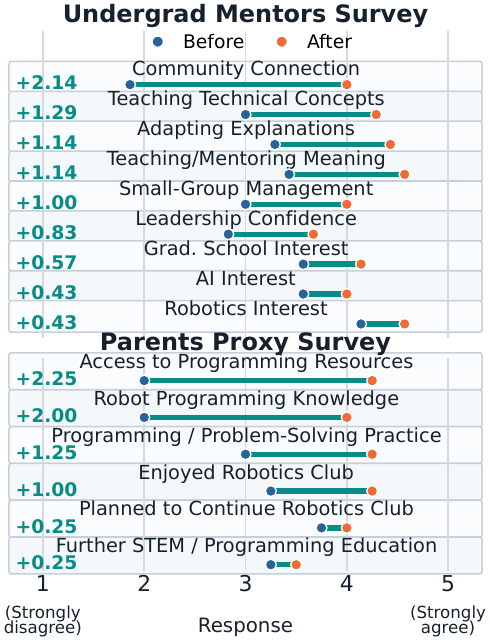}
    \caption{Mean retrospective pre-then-post survey responses from undergraduate mentors ($n=7$) and parents of participating FLL students ($n=4$). Responses used a five-point Likert scale from Strongly Disagree to Strongly Agree. One undergraduate did not provide a pre-course leadership response, giving $n=6$ for the paired leadership measure.}
    \label{fig:survey_res}
    \vspace{-10pt}
\end{figure}

\section{Related Work}

University--K--12 service learning and near-peer mentorship are well-established approaches in computing and engineering education. Credit-bearing programs have taught undergraduates to provide sustained community service and mentor robotics teams \cite{coyle2005epics,yamamoto2023cising,kafai2013cascading,karp2014pipeline,bhounsule2017robotics,matar2024cascade}. Rural STEM ecosystem research extends this partnership model through university coordination, co-design, local leadership, and community ownership \cite{nasem2025rural,kavanagh2022partnership,timko2023ruralecosystems,bhaduri2022codesigning,froehlich2026sustainability}. These fields motivate ARC's credit-bearing mentor course and development of mentoring capacity within participating schools.

ARC focuses on the rural mentorship gap created when schools may lie beyond practical travel distance of a university. Primary hubs 
support nearby teams, while mature K--12 programs can become secondary hubs that serve additional schools. Through this, mentorship can propagate beyond the fixed catchments of colleges through a self-reinforcing loop. This transfer of ownership is consistent with broader accounts of educational scale \cite{coburn2003scale,morel2019scale,nasem2025scaling}, while research on educational robotics, FIRST LEGO League, and AI literacy supports the workshop setting and instructional content ARC provides \cite{anwar2019educationalrobotics,graffin2022morethanrobots,long2020ailiteracy,touretzky2019year}.

\section{Trial Deployment}


This trial tested the fundamental components of ARC: a primary hub built around the ARC course and a recurring K--12 workshop. We did not attempt to test the full regional system, which would require several years for supported programs to mature and new hubs to form. Instead, the trial focused on whether undergraduate students could be quickly prepared as useful mentors, and whether a new robotics program could regularly attend and receive support through a primary hub. Establishing these components provides a foundation for larger deployments.

The trial was conducted at one large Indiana university located in a small urban area surrounded by substantial rural space. Three FLL teams from one school in a rural part of the surrounding county participated. This was the school's first year with an FLL program, and their teachers had previously been uncertain about starting the program because of limited access to mentorship. The course instructor had 12 years of prior experience mentoring competitive high-school robotics teams, as well as prior university teaching experience.

Nine undergraduate students enrolled in the ARC course. The course met once each week for a two-hour block. The first four weeks focused on preparing the undergraduates as mentors. Afterward, approximately 90 minutes were used for a workshop with the three teams, while the remaining 30 minutes were used for continuing course seminars. The teams traveled to the university for the workshops, where the undergraduate mentors worked with them using LEGO robots and two provided FLL game arenas.



We evaluated the trial using anonymous end-of-program surveys with pre-then-post questions -- respondents were asked after the program to rate their perceptions both before and after participation. This design can reduce response-shift effects when participation changes respondents' understanding of the construct being rated, since both ratings use the same post-program frame of reference. However, retrospective ratings can also be affected by recall and current-state biases, and should therefore be interpreted as retrospective self-assessments rather than direct longitudinal measurements.

Questions used a five-point Likert scale from Strongly Disagree to Strongly Agree, along with several post-program and open-response questions. To keep the evaluation limited to adult participants, K--12 outcomes were assessed through parent reports rather than direct surveys of children. Seven of the nine undergraduates and four parents completed the surveys. Survey details are available in Appendix~\ref{sec:survey}.

The undergraduate survey measured outcomes related to mentoring, community connection, leadership, and interest in AI, robotics, and graduate study. Figure~\ref{fig:survey_res} shows the pre-then-post results. Confidence teaching technical concepts increased from $3.00 \pm 1.15$ to $4.29 \pm 0.49$, while confidence adapting explanations increased from $3.29 \pm 1.11$ to $4.43 \pm 0.53$. Confidence managing small groups increased from $3.00 \pm 1.15$ to $4.00 \pm 0.58$. The largest change was in connection to the local community, which increased from $1.86 \pm 1.07$ to $4.00 \pm 0.82$. Finding teaching and mentoring enjoyable and meaningful increased from $3.43 \pm 0.79$ to $4.57 \pm 0.53$. Interest in AI, robotics, and graduate study also increased, although the changes were smaller.

Parents reported similar positive changes in several outcomes for their children. Ratings of basic robot programming knowledge increased from $2.00 \pm 1.41$ to $4.00 \pm 1.41$, while access to needed programming resources increased from $2.00 \pm 0.82$ to $4.25 \pm 0.50$. Opportunities to practice programming and problem solving increased from $3.00 \pm 0.82$ to $4.25 \pm 0.96$, and enjoyment of robotics club increased from $3.25 \pm 0.50$ to $4.25 \pm 1.50$. Changes in plans to continue robotics and interest in further STEM education were smaller. Parents also rated their children's experiences with the undergraduate mentors positively, with ratings of $4.75 \pm 0.50$ for enjoying work with the mentors, $4.50 \pm 0.58$ for feeling supported and encouraged, and $4.50 \pm 0.58$ for helping the team prepare for competition.

The three teams also competed at a 29-team regional FLL qualifier. During the competition, two reached the upper half of the field, including one that briefly reached the top quartile. Although their final standings were lower -- two in the third quartile and one near the top of the fourth -- this represents a strong first-season showing for three newly formed teams competing in their first tournament.

\section{Discussion \& Lessons Learned}

\noindent\textbf{Evidence from the trial.} The trial shows that undergraduates can be prepared within a short course to take on direct mentoring roles, and that new robotics programs can receive regular technical support through a primary hub. The undergraduate results also suggest benefits to mentor growth, particularly in teaching confidence and connection to local community. Parent responses indicate that the workshops provided useful programming resources and support to the participating teams. Given the small population, these results should be interpreted as initial evidence of feasibility and perceived benefit rather than precise estimates of program effects or a causal evaluation.

\noindent\textbf{Lessons for mentor preparation.} The trial identified two areas for improving mentor preparation: providing greater structure for novice mentors and better preparing them to manage participation within K--12 teams. Mentoring sessions were intentionally flexible because teams differed in pace, technical needs, and group dynamics, but five of six substantive improvement responses requested more structure, like clearer pacing, checkpoints, or milestones (Appendix~\ref{sec:survey_open}). Future deployments should pair flexible mentoring with clear milestone checklists and fallback activities, consistent with challenges reported in related service-learning programs \cite{bhounsule2017robotics,froehlich2026sustainability}. Mentors also sometimes struggled to maintain balanced participation when individual students dominated programming or robot work despite turn-taking and role rotation. Future preparation should include practical strategies for enforcing these structures respectfully, redirecting students, and coordinating with team coaches.

\noindent\textbf{Implications for scaling ARC.} The growth simulations suggest that ARC could scale substantially, with stronger assumptions producing progressively broader reach and more pronounced self-reinforcing growth, sometimes becoming superlinear. Structural assumptions make the model stronger evidence for growth shape than for exact long-horizon school counts. The ablations further show that the two hub mechanisms play complementary roles: primary hubs seed growth across broad regions, while secondary hubs allow mentorship to propagate recursively beyond them. Developing mature supported schools into durable secondary hubs should be an explicit objective of future ARC evaluations.

\noindent\textbf{Future work.} Future multi-site, multi-year deployments will expand the number of participating schools and primary hubs, follow supported programs as they mature, and observe the formation and persistence of secondary hubs. These observations will provide empirical estimates for the model's transition parameters, while working in real communities to close the rural robotics access gap nationwide.

\clearpage
\bibliography{aaai2027}

@article{williams2021robot,
  author  = {Williams, Randi and Kaputsos, Stephen P. and Breazeal, Cynthia},
  title   = {Teacher Perspectives on How To Train Your Robot: A Middle School AI and Ethics Curriculum},
  journal = {Proceedings of AAAI Conference on AI},
  volume  = {35},
  number  = {17},
  pages   = {15678--15686},
  year    = {2021},
  doi     = {10.1609/aaai.v35i17.17847},
  url     = {https://ojs.aaai.org/index.php/AAAI/article/view/17847}
}

@article{alvarez2022socially,
  author  = {Alvarez, Lauren and Gransbury, Isabella and Catet{\'e}, Veronica and Barnes, Tiffany and Led{\'e}czi, {\'A}kos and Grover, Shuchi},
  title   = {A Socially Relevant Focused AI Curriculum Designed for Female High School Students},
  journal = {Proceedings of AAAI Conference on AI},
  volume  = {36},
  number  = {11},
  pages   = {12698--12705},
  year    = {2022},
  doi     = {10.1609/aaai.v36i11.21546},
  url     = {https://ojs.aaai.org/index.php/AAAI/article/view/21546}
}

@article{krakowski2022authentic,
  author  = {Krakowski, Ari and Greenwald, Eric and Hurt, Timothy and Nonnecke, Brandie and Cannady, Matthew},
  title   = {Authentic Integration of Ethics and AI through Sociotechnical, Problem-Based Learning},
  journal = {Proceedings of AAAI Conference on AI},
  volume  = {36},
  number  = {11},
  pages   = {12774--12782},
  year    = {2022},
  doi     = {10.1609/aaai.v36i11.21556},
  url     = {https://ojs.aaai.org/index.php/AAAI/article/view/21556}
}

@article{stokes2023robotics,
  author  = {Stokes, Allison and Aurini, Janice and Rizk, Jessica and Gorbet, Rob and McLevey, John},
  title   = {Using Robotics to Support the Acquisition of {STEM} and 21st-Century Competencies: Promising (and Practical) Directions},
  journal = {Canadian Journal of Education / Revue canadienne de l'éducation},
  year    = {2023},
  volume  = {45},
  number  = {4},
  pages   = {1141--1170},
  doi     = {10.53967/cje-rce.5455}
}

@article{meador2025rural,
  author  = {Meador, Audrey and Raygoza, Ana and Irvin, Matthew and Starrett, Amber and Quiroz, Bianca and Cartiff, Brian},
  title   = {A Cross-Case Analysis of Rural Robotics Teacher Leaders' Identity Development},
  journal = {Frontiers in Education},
  year    = {2025},
  volume  = {10},
  pages   = {1578584},
  doi     = {10.3389/feduc.2025.1578584}
}

@article{coyle2005epics,
  author  = {Coyle, Edward J. and Jamieson, Leah H. and Oakes, William C.},
  title   = {{EPICS}: Engineering Projects in Community Service},
  journal = {International Journal of Engineering Education},
  volume  = {21},
  number  = {1},
  pages   = {139--150},
  year    = {2005}
}

@article{yamamoto2023cising,
  author    = {Yamamoto, Fujiko Robledo and Barker, Lecia and Voida, Amy},
  title     = {{CISing} Up Service Learning: A Systematic Review of Service Learning Experiences in Computer and Information Science},
  journal   = {ACM Transactions on Computing Education},
  volume    = {23},
  number    = {3},
  articleno = {37},
  numpages  = {56},
  year      = {2023},
  doi       = {10.1145/3610776}
}

@inproceedings{kafai2013cascading,
  author    = {Kafai, Yasmin B. and Griffin, Jean and Burke, Quinn and Slattery, Michelle and Fields, Deborah A. and Powell, Rita M. and Grab, Michele and Davidson, Susan B. and Sun, Joseph S.},
  title     = {A Cascading Mentoring Pedagogy in a {CS} Service Learning Course to Broaden Participation and Perceptions},
  booktitle = {Proceedings of the 44th ACM Technical Symposium on Computer Science Education},
  pages     = {101--106},
  publisher = {Association for Computing Machinery},
  year      = {2013},
  doi       = {10.1145/2445196.2445228}
}

@article{karp2014pipeline,
  author  = {Karp, Tanja and Gale, Richard and Tan, Mangnan and Burnham, Greg},
  title   = {Hosting a Pipeline of {K--12} Robotics Competitions at a College of Engineering: A Review of Benefits and Challenges},
  journal = {International Journal for Service Learning in Engineering, Humanitarian Engineering and Social Entrepreneurship},
  pages   = {406--423},
  year    = {2014},
  doi     = {10.24908/ijsle.v0i0.5560}
}

@inproceedings{bhounsule2017robotics,
  author    = {Bhounsule, Pranav and Chaney, Deborah and Claeys, Lorena and Manteufel, Randall D.},
  title     = {Robotics Service Learning for Improving Learning Outcomes and Increasing Community Engagement},
  booktitle = {Proceedings of the 2017 ASEE Gulf-Southwest Section Annual Conference},
  publisher = {American Society for Engineering Education},
  address   = {Dallas, Texas},
  year      = {2017},
  url       = {https://pab47.github.io/papers/ASEE17_service_learning.pdf}
}

@inproceedings{matar2024cascade,
  author    = {Matar, Sophia and Wimer, Bailey and Shuvo, M. I. R. and Mahmud, Saifuddin and Kim, Jong-Hoon and Novak, Elena and Borgerding, Lisa},
  title     = {Cascade Mentoring Experience to Engage High School Learners in {AI} and Robotics Through Project-Based Learning},
  booktitle = {Intelligent Human Computer Interaction},
  series    = {Lecture Notes in Computer Science},
  volume    = {14531},
  pages     = {279--294},
  publisher = {Springer},
  year      = {2024},
  doi       = {10.1007/978-3-031-53827-8_26}
}

@book{nasem2025rural,
  author    = {{National Academies of Sciences, Engineering, and Medicine}},
  title     = {{K--12 STEM} Education and Workforce Development in Rural Areas},
  publisher = {The National Academies Press},
  address   = {Washington, DC},
  year      = {2025},
  doi       = {10.17226/28269}
}

@article{kavanagh2022partnership,
  author  = {Kavanagh, Kathleen and DeWaters, Jan and Rivera, Seema and Richards, Melissa Carole and Ramsdell, Michael and Galluzzo, Ben},
  title   = {A University--Community Partnership Model to Support Rural {STEM} Teaching and Student Engagement},
  journal = {Theory \& Practice in Rural Education},
  volume  = {12},
  number  = {2},
  pages   = {229--248},
  year    = {2022},
  doi     = {10.3776/tpre.2022.v12n2p229-248}
}

@article{timko2023ruralecosystems,
  author  = {Timko, Gary and Harris, Maren and Hayde, Dolly and Peterman, Karen},
  title   = {Sustainable Development of Community-Supported {STEM}-Learning Ecosystems in Rural Areas of the United States},
  journal = {Community Development Journal},
  volume  = {58},
  number  = {3},
  pages   = {492--511},
  year    = {2023},
  doi     = {10.1093/cdj/bsac016}
}

@article{bhaduri2022codesigning,
  author  = {Bhaduri, Srinjita and Biddy, Quentin and Elliott, Colin Hennessy and Jacobs, Jennifer K. and Rummel, Melissa and Ristvey, John and Sumner, Tammy and Recker, Mimi},
  title   = {Co-designing a Rural Research--Practice Partnership to Design and Support {STEM} Pathways for Rural Youth},
  journal = {Theory \& Practice in Rural Education},
  volume  = {12},
  number  = {2},
  pages   = {45--70},
  year    = {2022},
  doi     = {10.3776/tpre.2022.v12n2p45-70}
}

@article{froehlich2026sustainability,
  author  = {Froehlich, Dominik Emanuel and Raberger, Julia},
  title   = {Sustainability-by-Design in School--University Partnerships: The Case of the Teaching Clinic},
  journal = {School-University Partnerships},
  volume  = {19},
  number  = {2},
  pages   = {273--287},
  year    = {2026},
  doi     = {10.1108/SUP-03-2026-0015}
}

@article{coburn2003scale,
  author  = {Coburn, Cynthia E.},
  title   = {Rethinking Scale: Moving Beyond Numbers to Deep and Lasting Change},
  journal = {Educational Researcher},
  volume  = {32},
  number  = {6},
  pages   = {3--12},
  year    = {2003},
  doi     = {10.3102/0013189X032006003}
}

@article{morel2019scale,
  author  = {Morel, Richard Paquin and Coburn, Cynthia E. and Catterson, Amy Koehler and Higgs, Jennifer},
  title   = {The Multiple Meanings of Scale: Implications for Researchers and Practitioners},
  journal = {Educational Researcher},
  volume  = {48},
  number  = {6},
  pages   = {369--377},
  year    = {2019},
  doi     = {10.3102/0013189X19860531}
}

@book{nasem2025scaling,
  author    = {{National Academies of Sciences, Engineering, and Medicine}},
  title     = {Scaling and Sustaining {Pre-K--12 STEM} Education Innovations: Systemic Challenges, Systemic Responses},
  publisher = {The National Academies Press},
  address   = {Washington, DC},
  year      = {2025},
  doi       = {10.17226/27950}
}

@article{anwar2019educationalrobotics,
  author  = {Anwar, Saira and Bascou, Nicholas A. and Menekse, Muhsin and Kardgar, Asefeh},
  title   = {A Systematic Review of Studies on Educational Robotics},
  journal = {Journal of Pre-College Engineering Education Research},
  volume  = {9},
  number  = {2},
  pages   = {19--42},
  year    = {2019},
  doi     = {10.7771/2157-9288.1223}
}

@article{graffin2022morethanrobots,
  author  = {Graffin, Michael and Sheffield, Rachel and Koul, Rekha},
  title   = {More than Robots: Reviewing the Impact of the {{FIRST LEGO League}} Challenge Robotics Competition on School Students' {STEM} Attitudes, Learning, and Twenty-First Century Skill Development},
  journal = {Journal for STEM Education Research},
  volume  = {5},
  number  = {3},
  pages   = {322--343},
  year    = {2022},
  doi     = {10.1007/s41979-022-00078-2}
}

@inproceedings{long2020ailiteracy,
  author    = {Long, Duri and Magerko, Brian},
  title     = {What Is {AI} Literacy? Competencies and Design Considerations},
  booktitle = {Proceedings of the 2020 CHI Conference on Human Factors in Computing Systems},
  pages     = {1--16},
  publisher = {Association for Computing Machinery},
  year      = {2020},
  doi       = {10.1145/3313831.3376727}
}

@article{touretzky2019year,
  author  = {Touretzky, David S. and Gardner-McCune, Christina and Breazeal, Cynthia and Martin, Fred and Seehorn, Deborah},
  title   = {A Year in {K--12 AI} Education},
  journal = {AI Magazine},
  volume  = {40},
  number  = {4},
  pages   = {88--90},
  year    = {2019},
  doi     = {10.1609/aimag.v40i4.5289}
}

@article{morris1991factorial,
  author  = {Morris, Max D.},
  title   = {Factorial Sampling Plans for Preliminary Computational Experiments},
  journal = {Technometrics},
  year    = {1991},
  volume  = {33},
  number  = {2},
  pages   = {161--174},
  doi     = {10.1080/00401706.1991.10484804}
}

@article{campolongo2007effective,
  author  = {Campolongo, Francesca and Cariboni, Jessica and Saltelli, Andrea},
  title   = {An Effective Screening Design for Sensitivity Analysis of Large Models},
  journal = {Environmental Modelling \& Software},
  year    = {2007},
  volume  = {22},
  number  = {10},
  pages   = {1509--1518},
  doi     = {10.1016/j.envsoft.2006.10.004}
}

\clearpage
\onecolumn
\appendix
\setcounter{secnumdepth}{1}
\makeatletter
\renewcommand{\@seccntformat}[1]{\csname the#1\endcsname.\quad}
\makeatother

\FloatBarrier
\section{Spatially-Explicit Markov Model of Growth Details}
\label{sec:model_full}

\subsection{Model Overview \& State Space}

The ARC growth model is a discrete-time stochastic model of individual K--12 school programs and potential college hubs. One time step represents one year. The state at year $t$ describes the system during that year, and one transition produces the complete state for year $t+1$. Geography is represented explicitly, so whether ARC can reach a school depends on the locations of the active hubs.

The unit of the school model is a school program. A school has an active program if at least one relevant FIRST team is registered to it. If a school has multiple teams, its program state is based on its most developed current team. Team count is not itself part of the state.

For school $i=1,\ldots,n$, where $n$ is the number of K--12 schools, let $F_i(t)\in\{N,E,M\}$ denote its FIRST program state, where $N$ is No program, $E$ is Early, and $M$ is Mature. Let $L_i(t)\in\{U,R,S,H\}$ denote its ARC state, where $U$ is Unreachable, $R$ is Reachable but not directly supported, $S$ is Supported, and $H$ is a secondary Hub. The complete state of school $i$ is $X_i(t)=(F_i(t),L_i(t))$.

Only nine combinations are valid. Table~\ref{tab:full_states} defines these school states.

\begin{table}[H]
\centering
\small
\begin{tabular}{@{}cll@{}}
\toprule
State & Program & ARC status \\
\midrule
$\mathrm{NU}$ & No program & Unreachable \\
$\mathrm{NR}$ & No program & Reachable \\
$\mathrm{EU}$ & Early & Unreachable \\
$\mathrm{ER}$ & Early & Reachable \\
$\mathrm{ES}$ & Early & Supported \\
$\mathrm{MU}$ & Mature & Unreachable \\
$\mathrm{MR}$ & Mature & Reachable \\
$\mathrm{MS}$ & Mature & Supported \\
$\mathrm{MH}$ & Mature & Secondary hub \\
\bottomrule
\end{tabular}
\caption{Valid K--12 school states in the full spatial model.}
\label{tab:full_states}
\end{table}

There are no $\mathrm{NS}$, $\mathrm{NH}$, or $\mathrm{EH}$ states. Supported schools must have an active program, and secondary hubs must have a Mature program. ARC assistance before program creation is represented by the $\mathrm{NR}$ state and the ARC launch probability defined below.

For potential primary-hub college $p=1,\ldots,m$, where $m$ is the number of potential primary-hub colleges, $C_p(t)\in\{0,1\}$ denotes whether the college is operating as an ARC primary hub. In particular, $C_p(t)=1$ means college $p$ is operating as a primary hub. The complete regional state is $Y(t)=\left(X_1(t),\ldots,X_n(t),C_1(t),\ldots,C_m(t)\right)$.

Several transition rules are fixed by the model. A school with no program can start only as Early, so there is no direct $N\rightarrow M$ transition. If a program closes, any later restart begins again as Early. Reachability alone does not change program maturation, program-closure, or program-backslide probabilities, so $\mathrm{EU}$ and $\mathrm{ER}$ use the same program dynamics, as do $\mathrm{MU}$ and $\mathrm{MR}$. Direct support may change these dynamics. A school can become a secondary hub only if it begins the year with a Mature program and remains Mature through the transition. Thus an Early program may mature in one step, but it cannot mature and become a hub in the same step. Programs may also undergo program closure or backslide from Mature to Early, Supported schools may lose support or reachability, and primary and secondary hubs may stop operating.

\subsection{Parameters and Calculated Quantities}

Table~\ref{tab:full_parameters} defines the parameters and inputs used by the full model. A superscript $B$ denotes unsupported program dynamics, $S$ denotes directly Supported program dynamics, and $H$ denotes secondary-hub program dynamics.

\begin{table*}[t]
\centering
\small
\begin{tabular}{@{}ll@{}}
\toprule
Symbol & Meaning \\
\midrule
$q_i^B(t)$
& Natural $N\rightarrow E$ start or restart probability for school $i$ \\

$\ell_i(t)$
& Probability that ARC launches an Early program at an $\mathrm{NR}$ school after no natural start \\

$\mu_E^B$
& Early $\rightarrow$ Mature probability for an unsupported program \\

$\delta_E^B$
& Program-closure probability for an unsupported Early program \\

$\mu_E^S$
& Early $\rightarrow$ Mature probability for a Supported school \\

$\delta_E^S$
& Program-closure probability for a Supported Early program \\

$\nu_M^B$
& Mature $\rightarrow$ Early program-backslide probability for an unsupported program \\

$\delta_M^B$
& Program-closure probability for an unsupported Mature program \\

$\nu_M^S$
& Mature $\rightarrow$ Early program-backslide probability for a Supported school \\

$\delta_M^S$
& Program-closure probability for a Supported Mature program \\

$\nu_M^H$
& Mature $\rightarrow$ Early program-backslide probability for a secondary hub \\

$\delta_M^H$
& Program-closure probability for a secondary hub \\

$\chi_i^B(t)$
& Probability that an unsupported school with a Mature program becomes a secondary hub \\

$\chi_i^S(t)$
& Probability that a Supported school with a Mature program becomes a secondary hub \\

$\rho_i(t)$
& Probability that an existing secondary hub remains a secondary hub if its program remains Mature \\

$\kappa_p(t)$
& Probability that inactive potential primary hub $p$ begins operating \\

$\xi_p(t)$
& Probability that active primary hub $p$ stops operating \\

$b_{hi}$
& Binary indicator that primary or secondary hub $h$ can geographically serve school $i$ \\

$w_i^{new}(t+1)$
& Probability that an active, reachable, non-hub school receives new direct ARC support \\

$w_i^{ret}(t+1)$
& Probability that an active, reachable, non-hub school retains existing direct ARC support \\
\bottomrule
\end{tabular}
\caption{Parameters and structural inputs of the full spatial model.}
\label{tab:full_parameters}
\end{table*}

The parameters in Table~\ref{tab:full_parameters} are combined into a small number of calculated quantities used in the transition probabilities. These calculations follow the model's annual update: first program-state changes, then the hub network and geographic reach, then direct ARC support.

For program-state changes, the probability that a program remains at its current development level is whatever probability remains after its possible forward or backward transitions:
\[
r_E^B=1-\delta_E^B-\mu_E^B,
\qquad
r_E^S=1-\delta_E^S-\mu_E^S,
\qquad
r_M^B=1-\delta_M^B-\nu_M^B,
\qquad
r_M^S=1-\delta_M^S-\nu_M^S,
\qquad
r_M^H=1-\delta_M^H-\nu_M^H.
\]

Essentially, each $r$ is the probability that the program stays at the same development level for another year. For an Early program, this means it neither undergoes program closure nor matures. For a Mature program, it means it neither undergoes program closure nor backslides to Early. For a secondary hub, $r_M^H$ refers to the probability that its program remains Mature. Whether the school also remains a secondary hub is handled separately by $\rho_i(t)$.

A Reachable school with no program can start either naturally or through ARC. ARC attempts an assisted program launch only when a natural start does not occur, so the total probability that an $\mathrm{NR}$ school enters an Early program is
\[
s_i(t)
=
q_i^B(t)
+
\left(1-q_i^B(t)\right)\ell_i(t).
\]
Its probability of remaining without a program is $1-s_i(t)=(1-q_i^B(t))(1-\ell_i(t))$.

The next part of the update determines the hub network and, from it, which schools are geographically reachable. For compact notation, let $J_i(t)=\mathbf{1}\{X_i(t)=\mathrm{MH}\}$, so $J_i(t)=1$ when school $i$ is a secondary hub and $0$ otherwise. The fixed quantity $b_{hi}$ equals one when hub $h$ can geographically serve school $i$ and zero otherwise, where $h$ may refer to either a primary or secondary hub. In the reachability calculation, $p$ indexes potential primary-hub colleges and $j$ indexes K--12 schools. Here, the $+$ superscript indicates reachability for year $t+1$ after the next-year hub states have been determined:
\[
g_i^+
=
1-
\prod_{p=1}^{m}
\left(1-b_{pi}C_p(t+1)\right)
\prod_{\substack{j=1\\j\neq i}}^{n}
\left(1-b_{ji}J_j(t+1)\right).
\]
Thus $g_i^+=1$ if at least one active primary hub or another active secondary hub can serve school $i$, and $g_i^+=0$ otherwise. The exclusion $j\neq i$ prevents a school from making itself reachable through its own hub status.

Finally, an active non-hub school that is reachable may receive direct ARC support. Its final ARC status is divided between Unreachable, Reachable but unsupported, and Supported according to
\[
1-g_i^+,
\qquad
g_i^+(1-w_i),
\qquad
g_i^+w_i,
\]
where $w_i=w_i^{ret}(t+1)$ if the school was already in a Supported state and $w_i=w_i^{new}(t+1)$ otherwise. A school with no program cannot be Supported, so a no-program destination is divided only between $\mathrm{NU}$ and $\mathrm{NR}$.

The model requires $\chi_i^S(t)\geq\chi_i^B(t)$, so a Supported school with a Mature program is at least as likely to become a secondary hub as an otherwise comparable unsupported school with a Mature program. It also requires $w_i^{ret}(t+1)\geq w_i^{new}(t+1)$, so retaining an existing direct-support relationship is at least as likely as creating a new one. All input probabilities lie in $[0,1]$, with
\[
\delta_E^B+\mu_E^B\leq1,
\qquad
\delta_E^S+\mu_E^S\leq1,
\qquad
\delta_M^B+\nu_M^B\leq1,
\qquad
\delta_M^S+\nu_M^S\leq1,
\qquad
\delta_M^H+\nu_M^H\leq1.
\]
These constraints ensure that the corresponding $r$ probabilities are nonnegative.

The quantities $\ell_i(t)$, $w_i^{new}(t+1)$, and $w_i^{ret}(t+1)$ describe the chances that ARC launches a new program, begins direct support, or continues direct support at a school. A particular implementation may specify these probabilities directly or determine them from explicit limits on how many schools can be launched or supported. If only a fixed number of launch or support spaces are available, the implementation must also specify how the schools receiving those spaces are selected.

\subsection{Annual Transition Process}

Each $t\rightarrow t+1$ update produces one new state for every school and college. The calculation proceeds in five steps.

\begin{enumerate}
    \item Each school receives a program outcome: No program, Early, or Mature. The probabilities depend on its state at year $t$. An $\mathrm{NR}$ school additionally has the ARC launch opportunity $\ell_i(t)$.

    \item Potential primary hubs update using $\kappa_p(t)$ and $\xi_p(t)$. Schools that began the year with a Mature program and remain Mature may become secondary hubs using $\chi_i^B(t)$ or $\chi_i^S(t)$. Existing secondary hubs whose programs remain Mature remain secondary hubs with probability $\rho_i(t)$.

    \item The resulting primary and secondary-hub network is used to calculate $g_i^+$ for every non-hub school.

    \item Active reachable non-hub schools receive or retain direct ARC support according to $w_i^{new}(t+1)$ or $w_i^{ret}(t+1)$.

    \item The program outcome and ARC outcome are combined into one of the nine valid states in Table~\ref{tab:full_states}.
\end{enumerate}

These are dependencies within one annual transition. For example, a school that begins the year with an Early program cannot become a hub at that boundary even if its program becomes Mature.

For each transition, the source is the school's state at year $t$ and the destination is its state at year $t+1$. A school transition probability contains a program factor, an optional hub factor, and a final reach/support factor. Table~\ref{tab:school_transitions} gives all 70 legal school transitions. Any source--destination pair not listed has probability zero.

\begin{table*}[t]
\centering
\scriptsize

\begin{minipage}[t]{0.32\textwidth}
\centering
\textbf{$\mathrm{NU}$}\\[2pt]
\begin{tabular}{@{}cl@{}}
\toprule
To & Probability \\
\midrule
$\mathrm{NU}$ & $(1-q_i^B)(1-g_i^+)$ \\
$\mathrm{NR}$ & $(1-q_i^B)g_i^+$ \\
$\mathrm{EU}$ & $q_i^B(1-g_i^+)$ \\
$\mathrm{ER}$ & $q_i^B g_i^+(1-w_i^{new})$ \\
$\mathrm{ES}$ & $q_i^B g_i^+w_i^{new}$ \\
\bottomrule
\end{tabular}
\end{minipage}
\hfill
\begin{minipage}[t]{0.32\textwidth}
\centering
\textbf{$\mathrm{NR}$}\\[2pt]
\begin{tabular}{@{}cl@{}}
\toprule
To & Probability \\
\midrule
$\mathrm{NU}$ & $(1-q_i^B)(1-\ell_i)(1-g_i^+)$ \\
$\mathrm{NR}$ & $(1-q_i^B)(1-\ell_i)g_i^+$ \\
$\mathrm{EU}$ & $s_i(1-g_i^+)$ \\
$\mathrm{ER}$ & $s_i g_i^+(1-w_i^{new})$ \\
$\mathrm{ES}$ & $s_i g_i^+w_i^{new}$ \\
\bottomrule
\end{tabular}
\end{minipage}
\hfill
\begin{minipage}[t]{0.32\textwidth}
\centering
\textbf{$\mathrm{EU}$}\\[2pt]
\begin{tabular}{@{}cl@{}}
\toprule
To & Probability \\
\midrule
$\mathrm{NU}$ & $\delta_E^B(1-g_i^+)$ \\
$\mathrm{NR}$ & $\delta_E^B g_i^+$ \\
$\mathrm{EU}$ & $r_E^B(1-g_i^+)$ \\
$\mathrm{ER}$ & $r_E^B g_i^+(1-w_i^{new})$ \\
$\mathrm{ES}$ & $r_E^B g_i^+w_i^{new}$ \\
$\mathrm{MU}$ & $\mu_E^B(1-g_i^+)$ \\
$\mathrm{MR}$ & $\mu_E^B g_i^+(1-w_i^{new})$ \\
$\mathrm{MS}$ & $\mu_E^B g_i^+w_i^{new}$ \\
\bottomrule
\end{tabular}
\end{minipage}

\vspace{8pt}

\begin{minipage}[t]{0.32\textwidth}
\centering
\textbf{$\mathrm{ER}$}\\[2pt]
\begin{tabular}{@{}cl@{}}
\toprule
To & Probability \\
\midrule
$\mathrm{NU}$ & $\delta_E^B(1-g_i^+)$ \\
$\mathrm{NR}$ & $\delta_E^B g_i^+$ \\
$\mathrm{EU}$ & $r_E^B(1-g_i^+)$ \\
$\mathrm{ER}$ & $r_E^B g_i^+(1-w_i^{new})$ \\
$\mathrm{ES}$ & $r_E^B g_i^+w_i^{new}$ \\
$\mathrm{MU}$ & $\mu_E^B(1-g_i^+)$ \\
$\mathrm{MR}$ & $\mu_E^B g_i^+(1-w_i^{new})$ \\
$\mathrm{MS}$ & $\mu_E^B g_i^+w_i^{new}$ \\
\bottomrule
\end{tabular}
\end{minipage}
\hfill
\begin{minipage}[t]{0.32\textwidth}
\centering
\textbf{$\mathrm{ES}$}\\[2pt]
\begin{tabular}{@{}cl@{}}
\toprule
To & Probability \\
\midrule
$\mathrm{NU}$ & $\delta_E^S(1-g_i^+)$ \\
$\mathrm{NR}$ & $\delta_E^S g_i^+$ \\
$\mathrm{EU}$ & $r_E^S(1-g_i^+)$ \\
$\mathrm{ER}$ & $r_E^S g_i^+(1-w_i^{ret})$ \\
$\mathrm{ES}$ & $r_E^S g_i^+w_i^{ret}$ \\
$\mathrm{MU}$ & $\mu_E^S(1-g_i^+)$ \\
$\mathrm{MR}$ & $\mu_E^S g_i^+(1-w_i^{ret})$ \\
$\mathrm{MS}$ & $\mu_E^S g_i^+w_i^{ret}$ \\
\bottomrule
\end{tabular}
\end{minipage}
\hfill
\begin{minipage}[t]{0.32\textwidth}
\centering
\textbf{$\mathrm{MU}$}\\[2pt]
\begin{tabular}{@{}cl@{}}
\toprule
To & Probability \\
\midrule
$\mathrm{NU}$ & $\delta_M^B(1-g_i^+)$ \\
$\mathrm{NR}$ & $\delta_M^B g_i^+$ \\
$\mathrm{EU}$ & $\nu_M^B(1-g_i^+)$ \\
$\mathrm{ER}$ & $\nu_M^B g_i^+(1-w_i^{new})$ \\
$\mathrm{ES}$ & $\nu_M^B g_i^+w_i^{new}$ \\
$\mathrm{MU}$ & $r_M^B(1-\chi_i^B)(1-g_i^+)$ \\
$\mathrm{MR}$ & $r_M^B(1-\chi_i^B)g_i^+(1-w_i^{new})$ \\
$\mathrm{MS}$ & $r_M^B(1-\chi_i^B)g_i^+w_i^{new}$ \\
$\mathrm{MH}$ & $r_M^B\chi_i^B$ \\
\bottomrule
\end{tabular}
\end{minipage}

\vspace{8pt}

\begin{minipage}[t]{0.32\textwidth}
\centering
\textbf{$\mathrm{MR}$}\\[2pt]
\begin{tabular}{@{}cl@{}}
\toprule
To & Probability \\
\midrule
$\mathrm{NU}$ & $\delta_M^B(1-g_i^+)$ \\
$\mathrm{NR}$ & $\delta_M^B g_i^+$ \\
$\mathrm{EU}$ & $\nu_M^B(1-g_i^+)$ \\
$\mathrm{ER}$ & $\nu_M^B g_i^+(1-w_i^{new})$ \\
$\mathrm{ES}$ & $\nu_M^B g_i^+w_i^{new}$ \\
$\mathrm{MU}$ & $r_M^B(1-\chi_i^B)(1-g_i^+)$ \\
$\mathrm{MR}$ & $r_M^B(1-\chi_i^B)g_i^+(1-w_i^{new})$ \\
$\mathrm{MS}$ & $r_M^B(1-\chi_i^B)g_i^+w_i^{new}$ \\
$\mathrm{MH}$ & $r_M^B\chi_i^B$ \\
\bottomrule
\end{tabular}
\end{minipage}
\hfill
\begin{minipage}[t]{0.32\textwidth}
\centering
\textbf{$\mathrm{MS}$}\\[2pt]
\begin{tabular}{@{}cl@{}}
\toprule
To & Probability \\
\midrule
$\mathrm{NU}$ & $\delta_M^S(1-g_i^+)$ \\
$\mathrm{NR}$ & $\delta_M^S g_i^+$ \\
$\mathrm{EU}$ & $\nu_M^S(1-g_i^+)$ \\
$\mathrm{ER}$ & $\nu_M^S g_i^+(1-w_i^{ret})$ \\
$\mathrm{ES}$ & $\nu_M^S g_i^+w_i^{ret}$ \\
$\mathrm{MU}$ & $r_M^S(1-\chi_i^S)(1-g_i^+)$ \\
$\mathrm{MR}$ & $r_M^S(1-\chi_i^S)g_i^+(1-w_i^{ret})$ \\
$\mathrm{MS}$ & $r_M^S(1-\chi_i^S)g_i^+w_i^{ret}$ \\
$\mathrm{MH}$ & $r_M^S\chi_i^S$ \\
\bottomrule
\end{tabular}
\end{minipage}
\hfill
\begin{minipage}[t]{0.32\textwidth}
\centering
\textbf{$\mathrm{MH}$}\\[2pt]
\begin{tabular}{@{}cl@{}}
\toprule
To & Probability \\
\midrule
$\mathrm{NU}$ & $\delta_M^H(1-g_i^+)$ \\
$\mathrm{NR}$ & $\delta_M^H g_i^+$ \\
$\mathrm{EU}$ & $\nu_M^H(1-g_i^+)$ \\
$\mathrm{ER}$ & $\nu_M^H g_i^+(1-w_i^{new})$ \\
$\mathrm{ES}$ & $\nu_M^H g_i^+w_i^{new}$ \\
$\mathrm{MU}$ & $r_M^H(1-\rho_i)(1-g_i^+)$ \\
$\mathrm{MR}$ & $r_M^H(1-\rho_i)g_i^+(1-w_i^{new})$ \\
$\mathrm{MS}$ & $r_M^H(1-\rho_i)g_i^+w_i^{new}$ \\
$\mathrm{MH}$ & $r_M^H\rho_i$ \\
\bottomrule
\end{tabular}
\end{minipage}

\caption{Complete school transition probabilities. Each small table is labeled by its source state and gives every legal destination from that state. Time arguments are omitted for readability.}
\label{tab:school_transitions}
\end{table*}

For each starting state, the probabilities of all possible next states sum to one. For an Early program, the possible program outcomes are program closure, remaining Early, or maturation. For a Mature program, they are program closure, backsliding to Early, or remaining Mature. If an unsupported or Supported school with a Mature program remains Mature, its probability is further divided between becoming a secondary hub and remaining a non-hub school. If an existing secondary hub remains Mature, its probability is further divided between remaining a secondary hub and ceasing to operate as one. Each active non-hub branch is then divided between the three ARC outcomes because
\[
(1-g_i^+)+g_i^+(1-w_i)+g_i^+w_i=1.
\]

Potential primary hubs have four transitions:
\[
\Pr(C_p(t+1)=1\mid C_p(t)=0)=\kappa_p(t),
\qquad
\Pr(C_p(t+1)=0\mid C_p(t)=0)=1-\kappa_p(t),
\]
\[
\Pr(C_p(t+1)=0\mid C_p(t)=1)=\xi_p(t),
\qquad
\Pr(C_p(t+1)=1\mid C_p(t)=1)=1-\xi_p(t).
\]

By default, the random program, hub, and support events for different schools and colleges are drawn independently once the current system state and their required inputs are fixed. The schools are nevertheless linked through geography because realized hub outcomes determine the reachability of other schools. If a concrete implementation has a fixed number of launch or support spaces, it must also specify how the schools receiving those spaces are selected.

\subsection{Initialization and Model Outputs}

A full spatial-model run requires the set of schools, an initial valid state $X_i(0)$ for each school, the set of potential primary-hub colleges, an initial value $C_p(0)$ for each college, the geographic relation $b_{hi}$, the model parameters, and any rules used to select schools when launch or support capacity is limited. The model does not assume a universal initial distribution. These inputs are properties of the region being modeled.

For any school state $s$, let $Z_s(t)=\sum_{i=1}^{n}\mathbf{1}\{X_i(t)=s\}$ denote the number of schools in that state.

The total number of schools with active FIRST programs is
\[
P(t)
=
Z_{\mathrm{EU}}(t)+Z_{\mathrm{ER}}(t)+Z_{\mathrm{ES}}(t)
+Z_{\mathrm{MU}}(t)+Z_{\mathrm{MR}}(t)+Z_{\mathrm{MS}}(t)
+Z_{\mathrm{MH}}(t).
\]

The total number of schools within the ARC network, including Reachable, Supported, and secondary-hub schools, is
\[
A(t)
=
Z_{\mathrm{NR}}(t)+Z_{\mathrm{ER}}(t)+Z_{\mathrm{ES}}(t)
+Z_{\mathrm{MR}}(t)+Z_{\mathrm{MS}}(t)+Z_{\mathrm{MH}}(t).
\]

The number of directly Supported schools is $S(t)=Z_{\mathrm{ES}}(t)+Z_{\mathrm{MS}}(t)$, the number of secondary hubs is $H(t)=Z_{\mathrm{MH}}(t)$, and the number of active primary hubs is $C(t)=\sum_{p=1}^{m}C_p(t)$.

\FloatBarrier
\section{Mean Field Approximation Details}
\label{sec:model_mf}

The mean-field model is an approximation of the spatially-explicit model in Appendix~\ref{sec:model_full}. It uses the same nine school states, the same two primary-hub college states, and the same allowed transitions. The difference is that individual schools and their exact geographic relationships are replaced by regional averages. Instead of sampling a state for every school, the model tracks the expected number of schools in each state.

Let $Z_s(t)$ denote the expected number of schools in school state $s$ at year $t$, where $s\in\{\mathrm{NU},\mathrm{NR},\mathrm{EU},\mathrm{ER},\mathrm{ES},\mathrm{MU},\mathrm{MR},\mathrm{MS},\mathrm{MH}\}$. These expected counts satisfy $\sum_s Z_s(t)=n$. Because they are expectations, they may be fractional. For example, $Z_{\mathrm{MH}}(t)=55.6$ means that the model predicts an average of 55.6 secondary hubs across comparable realizations, not that a physical system contains a fraction of a hub.

Similarly, let $C_0(t)$ and $C_1(t)$ denote the expected numbers of inactive and active potential primary hubs, with $C_0(t)+C_1(t)=m$.

\subsection{Mean-Field Replacements}

Most parameters from the full model are unchanged. The program maturation, program-closure, and program-backslide probabilities $\mu$, $\delta$, $\nu$, and their calculated retention probabilities $r$ retain the same definitions. The mean-field approximation replaces school-specific and geographically-specific quantities with the regional quantities shown in Table~\ref{tab:mf_replacements}.

\begin{table}[H]
\centering
\small
\begin{tabular}{@{}lll@{}}
\toprule
Full model & Mean field & Meaning \\
\midrule
$q_i^B(t)$ & $q^B$ & Regional natural start probability \\
$\ell_i(t)$ & $\bar{\ell}_t$ & Regional ARC launch probability \\
$\chi_i^B(t)$ & $\chi^B$ & Unsupported secondary-hub formation probability \\
$\chi_i^S(t)$ & $\chi^S$ & Supported secondary-hub formation probability \\
$\rho_i(t)$ & $\rho$ & Secondary-hub retention probability \\
$\kappa_p(t)$ & $\kappa$ & Primary-hub activation probability \\
$\xi_p(t)$ & $\xi$ & Primary-hub stopping probability \\
$b_{pi}$ & $c_C$ & Average coverage from one primary hub \\
$b_{ji}$ & $c_H$ & Average coverage from one secondary hub \\
$g_i^+$ & $\bar{g}_{t+1}$ & Regional probability of being reachable \\
$w_i^{new}$ & $\bar{w}^{new}$ & New direct-support probability \\
$w_i^{ret}$ & $\bar{w}^{ret}$ & Direct-support retention probability \\
\bottomrule
\end{tabular}
\caption{Quantities replaced by regional averages in the mean-field approximation.}
\label{tab:mf_replacements}
\end{table}

The mean-field model also uses $L$, the expected maximum number of successful new school-program launches that one active hub can produce in a year, and $K$, the expected number of schools that one active hub can directly support.

\subsection{Mean-Field Annual Update}

The annual update follows the same order as the full model. Program states change first, followed by hub formation, reachability, direct support, and the final school-state update.

A Reachable school with no program can still start naturally with probability $q^B$. ARC can attempt to launch a program among the remaining $\mathrm{NR}$ schools. The expected number of such schools available for an ARC launch is $(1-q^B)Z_{\mathrm{NR}}(t)$. The hubs active during year $t$ can together produce up to $L[C_1(t)+Z_{\mathrm{MH}}(t)]$ expected launches. The resulting ARC launch probability is therefore
\[
\bar{\ell}_t
=
\min\left(
1,
\frac{L[C_1(t)+Z_{\mathrm{MH}}(t)]}
{(1-q^B)Z_{\mathrm{NR}}(t)}
\right).
\]
If the denominator is zero, $\bar{\ell}_t=0$. As in the full model, launches during year $t$ use hubs that are already active during that year.

Expected primary-hub counts update as
\[
C_1(t+1)=C_1(t)(1-\xi)+C_0(t)\kappa,
\qquad
C_0(t+1)=C_0(t)(1-\kappa)+C_1(t)\xi.
\]

The expected number of secondary hubs at year $t+1$ is
\[
Z_{\mathrm{MH}}(t+1)
=
Z_{\mathrm{MH}}(t)r_M^H\rho
+
[Z_{\mathrm{MU}}(t)+Z_{\mathrm{MR}}(t)]r_M^B\chi^B
+
Z_{\mathrm{MS}}(t)r_M^S\chi^S.
\]
The three terms represent existing secondary hubs that remain Mature and continue operating, unsupported schools with Mature programs that become secondary hubs, and Supported schools with Mature programs that become secondary hubs. As in the full model, a school that begins the year with an Early program cannot become a secondary hub during the same transition.

Exact geographic coverage is then replaced by average coverage. Let $c_C$ be the probability that one active primary hub covers a representative school and $c_H$ the corresponding probability for one secondary hub. Mean-field reachability is
\[
\bar{g}_{t+1}
=
1-
(1-c_C)^{C_1(t+1)}
(1-c_H)^{Z_{\mathrm{MH}}(t+1)}.
\]
This is the main spatial approximation in the model. The full model determines whether each specific school is inside the service area of each specific hub. The mean-field model instead spreads each hub's expected coverage across the whole region. It therefore does not preserve geographic clustering or overlapping service areas. In particular, it can make geographic reach expand faster than the full spatial model when many real hubs would cover the same schools or remain concentrated in one part of the region.

Direct support then uses $\bar{w}^{new}$ for schools that were not previously Supported and $\bar{w}^{ret}$ for schools that were already in a Supported state. Let $A_{new}$ and $A_{ret}$ denote the corresponding expected masses of active, reachable, non-hub schools. The expected number of direct-support assignments is
\[
D_{\mathrm{support}}
=
\bar{w}^{new}A_{new}
+
\bar{w}^{ret}A_{ret}.
\]
The expected available support capacity is
\[
Q_{\mathrm{support}}
=
K[C_1(t+1)+Z_{\mathrm{MH}}(t+1)].
\]
The selected support probabilities are checked against $D_{\mathrm{support}}\leq Q_{\mathrm{support}}$. This capacity check does not change the transition probabilities automatically.

\subsection{State-Mass Update}

The mean-field model uses the same 70 legal school transitions listed in Table~\ref{tab:school_transitions}. Their probabilities are obtained by replacing the school-specific quantities with the corresponding mean-field quantities in Table~\ref{tab:mf_replacements}. For example, $\Pr(\mathrm{MS}\rightarrow\mathrm{MH})=r_M^S\chi^S$, and $\Pr(\mathrm{MS}\rightarrow\mathrm{MR})=r_M^S(1-\chi^S)\bar{g}_{t+1}(1-\bar{w}^{ret})$.

Let $\bar{p}_{s\rightarrow v}(t)$ denote the resulting mean-field transition probability from state $s$ to state $v$. The expected number of schools in destination state $v$ is then
\[
Z_v(t+1)
=
\sum_s Z_s(t)\bar{p}_{s\rightarrow v}(t),
\]
where the sum is over all nine school states. Because the transition probabilities from each starting state sum to one, the total expected school count remains $\sum_s Z_s(t)=n$ at every year.

A mean-field run therefore begins with expected counts for the nine school states and the two primary-hub college states. After initialization, the equations above deterministically produce the expected state counts for each subsequent year. Unlike the full spatial model, the mean-field model does not sample individual schools or produce realization-to-realization variation.
\FloatBarrier
\section{Spatially-Explicit Model Analysis of Indiana}
\label{sec:model_analysis}

\subsection{Quantities Held Fixed Across Scenarios}
All 12 Indiana optimism levels use the same 2025--2026 public K--12 school universe and the same natural FIRST program dynamics. Private K--12 schools are excluded from the modeling denominator to simplify data collection. The potential-college-hub population is not restricted to public institutions. A school is counted as having a FIRST program when at least one FIRST LEGO League Challenge, FIRST Tech Challenge, or FIRST Robotics Competition team is linked to that school. Program maturity is the maturity of the school's most-developed current team.

\begin{table}[H]
\centering
\small
\begin{tabularx}{\linewidth}{@{}l r X X@{}}
\toprule
Quantity & Fixed value & Rationale / calculation & Source or basis \\
\midrule
$n$ & $1{,}925$ & Indiana public schools serving at least one K--12 grade in the 2025--2026 directory; PK-only sites are excluded. & Indiana Department of Education, 2025--2026 Indiana School Directory.\textsuperscript{a} \\
$m$ & $32$ & Conventional Indiana campuses reporting at least three bachelor's-level completions in IPEDS CIP family 11 during 2023--2024. Public and private institutions are included; online and extension-only institutions are excluded. & IPEDS 2023--2024 completions and institutional locations.\textsuperscript{b} \\
$W$ & $3$ seasons & A team is Mature after the same team identity is linked to the same school in three consecutive FIRST seasons. A school is Mature if at least one of its current teams is Mature. A closure resets this history, implementing the model's no-memory restart assumption. & Fixed operational definition for mapping observed FIRST records into model states. \\
$q^B$ & $0.01229$ & Natural $N\to E$ start probability. Pooled 2021--2025 estimate: $(89-1)/7162=0.012287$, subtracting the one known 2025 ARC-created school start. & FIRST Team \& Event Search linked to IDOE public schools.\textsuperscript{c} \\
$\mu_E^B$ & $0.28947$ & Natural $E\to M$ maturation probability: $55/190$ Early school-years. & Same linked 2021--2025 public-school panel. \\
$\delta_E^B$ & $0.23684$ & Natural Early-program closure probability: $45/190$ Early school-years. & Same linked 2021--2025 public-school panel. \\
$\nu_M^B$ & $0.00575$ & Natural $M\to E$ backslide probability: $2/348$ Mature school-years. This is observable because maturity follows the most-developed team's continuous tenure at that school. & Same linked 2021--2025 public-school panel. \\
$\delta_M^B$ & $0.09770$ & Natural Mature-program closure probability: $34/348$ Mature school-years. & Same linked 2021--2025 public-school panel. \\
$p_{\mathrm{req}}^{\mathrm{ret}}$ & $1$ & Every previously Supported school that remains an eligible reachable active non-hub school requests continued support and receives priority over new support requests. Actual assignment remains subject to available capacity at a covering hub. & ARC operating assumption supplied from the trial. \\
$N(0)$ & $1{,}790$ schools & No accepted FIRST program link in 2025. & Conservative FIRST--IDOE school linkage. \\
$E(0)$ & $44$ schools & At least one accepted current FIRST program link, but no current team at that school has met the three-season maturity rule. & Conservative FIRST--IDOE school linkage. \\
$M(0)$ & $91$ schools & At least one current team identity is linked to that school in each of the final three consecutive seasons. & Conservative FIRST--IDOE school linkage. \\
$C_1(0)$ & $1$ college & One active primary hub at the start of the 2025--2026 analysis. & ARC trial fact. \\
$C_0(0)$ & $31$ colleges & Remaining potential primary hubs: $m-C_1(0)=32-1$. & Derived from $m$ and the trial initial condition. \\
$M_H(0)$ & $0$ schools & No secondary school hubs at the initial boundary. & ARC trial fact. \\
\bottomrule
\end{tabularx}

\vspace{0.4em}
\begin{minipage}{\linewidth}
\footnotesize
\textsuperscript{a} Indiana Department of Education, Indiana School Directory:
\url{https://www.in.gov/doe/it/data-center-and-reports/}

\textsuperscript{b} Integrated Postsecondary Education Data System (IPEDS), National Center for Education Statistics:
\url{https://nces.ed.gov/ipeds/}

\textsuperscript{c} FIRST Team \& Event Search:
\url{https://www.firstinspires.org/team-event-search}
\end{minipage}
\end{table}

The natural transition estimates pool the four post-COVID year-to-year transitions 2021--2022 through 2024--2025. FIRST organization records are linked conservatively to current IDOE public-school names using normalized school-name and locality information; ambiguous multi-school, community, home-school, and non-public organizations are not silently assigned to a public school. The resulting 2025 linked count of 135 public schools with at least one FIRST program should therefore be read as a conservative school-linked count, not as a claim that every unmatched FIRST organization is non-public.

\subsection{Quantities in Scenarios}

Parameters describing ARC effectiveness cannot yet be estimated reliably from the single-year trial. We therefore evaluate 12 coordinatewise ordered optimism levels. Level 1 uses the least favorable assumptions considered, Level 6 provides the moderate anchor used for the main comparison, and Level 12 uses the most favorable assumptions considered. Levels 2--5 interpolate linearly between Levels 1 and 6, while Levels 7--11 interpolate linearly between Levels 6 and 12. These levels are not confidence intervals or fitted estimates. Natural program dynamics and the underlying school and college populations remain fixed across all 12 levels.

\begin{table}[H]
\centering
\small
\begin{tabularx}{\linewidth}{@{}l r r r X@{}}
\toprule
Quantity & Level 1 & Level 6 & Level 12 & Interpretation / construction \\
\midrule
Reach radius & $15$ mi & $20$ mi & $25$ mi & Maximum great-circle distance over which an active hub can make ARC resources reachable to a school. \\

$L$ & $3$ & $3$ & $3$ & Maximum number of successful new school-program launches allocated to each hub active at the start of a year. Held fixed so stronger levels do not gain from greater per-hub launch throughput. \\

$K$ & $3$ & $6$ & $10$ & Expected number of direct-support slots per active hub. Three distinct schools are operationally feasible with demonstrated trial resources; 6--10 represents the estimated scalable range. \\

$p_{\mathrm{req}}^{\mathrm{new}}$ & $0.05$ & $0.10$ & $0.15$ & Annual probability that an eligible active, reachable, non-supported school requests new direct ARC support. Whether the request is filled depends on local hub capacity. \\

$\mu_E^S$ & $0.28947$ & $0.36184$ & $0.43421$ & Supported Early-program maturation. The three anchors use $1.0$, $1.25$, and $1.5$ times the observed natural maturation rate. \\

$\delta_E^S$ & $0.21316$ & $0.16579$ & $0.11842$ & Supported Early-program closure. The natural closure rate is multiplied by $0.9$, $0.7$, and $0.5$. \\

$\nu_M^S$ & $0.00517$ & $0.00402$ & $0.00287$ & Supported Mature-program backslide, using the same $0.9$, $0.7$, and $0.5$ multipliers. \\

$\delta_M^S$ & $0.08793$ & $0.06839$ & $0.04885$ & Supported Mature-program closure, using the same multipliers. \\

$\nu_M^H$ & $0.00172$ & $0.00115$ & $0.00057$ & Hub-program backslide, using $0.30$, $0.20$, and $0.10$ times the observed natural Mature backslide rate. \\

$\delta_M^H$ & $0.02931$ & $0.01954$ & $0.00977$ & Hub-program closure, using $0.30$, $0.20$, and $0.10$ times the observed natural Mature closure rate. \\

$\chi^B$ & $0.00125$ & $0.0030$ & $0.0035$ & Annual probability that an eligible Mature, non-supported program becomes a secondary hub. \\

$\chi^S$ & $0.010$ & $0.025$ & $0.030$ & Annual hub-formation probability for a Supported Mature program. Support is assumed to make deliberate hub training substantially more likely. \\

$\rho$ & $0.96$ & $0.97$ & $0.98$ & Probability that a secondary hub continues hosting, conditional on its program remaining Mature. \\

$\kappa$ & $0.012$ & $0.017$ & $0.020$ & Primary-hub activation probability. With 31 initially inactive colleges, these correspond to approximately $0.37$, $0.53$, and $0.62$ expected new primary hubs in the first year. \\

$\xi$ & $0.05$ & $0.045$ & $0.040$ & Primary-hub cessation probability, corresponding to mean uninterrupted active periods of approximately $20$, $22$, and $25$ years. \\
\bottomrule
\end{tabularx}
\end{table}

Geographic coverage in the full model is calculated directly from the coordinates of individual schools and hubs. School coordinates come primarily from NCES EDGE 2024--2025, with documented Census and NCES-derived fallback geocodes for unmatched directory entries; college coordinates come from IPEDS. At each optimism level, the reach radius defines a fixed hub-to-school coverage relation, and a school is reachable only when at least one hub that covers it is active. The mean-field coverage fractions $c_C$ and $c_H$ in Appendix~B are derived from the same radii but are not inputs to the spatial simulation.

The Supported-program rates are constructed from the empirically estimated natural rates. Level 1 assumes no improvement in maturation and only a small reduction in failure, while Levels 6 and 12 progressively increase maturation and reduce closure and backslide. Hub persistence is represented by both program stability and continued hosting. At Levels 1, 6, and 12, the resulting one-year probabilities that an active secondary hub remains a hub are approximately $0.930$, $0.950$, and $0.970$, corresponding under constant rates to expected uninterrupted active-hub durations of approximately $14.3$, $20.0$, and $33.2$ years.

The remaining quantities describe ARC operations and replication and therefore cannot yet be estimated from historical natural-growth data. In particular, $\chi^B$ and $\chi^S$ control secondary-hub formation, with $\chi^S>\chi^B$ representing the hypothesis that a school already receiving direct ARC support is more likely to be trained into a hub. $L$ controls successful program creation and is held fixed at three at all levels, while $K$ governs local support capacity and $p_{\mathrm{req}}^{\mathrm{new}}$ controls how often eligible schools request new direct support.

The spatial simulation enforces both launch and support capacity locally. Reachable No Program schools that do not start naturally are randomly ordered and assigned to their nearest covering hub with an available launch slot, using hubs active at the start of the year. After primary- and secondary-hub transitions determine the next-year network, continuing Supported schools receive priority for direct-support capacity.
New support requesters are then randomly ordered and assigned to their nearest covering hub with an available slot. Thus the $p_{\mathrm{req}}$ quantities govern requests, while the $w_i$ quantities give the resulting support-assignment probabilities.
When an intermediate optimism level gives a fractional value of $K$, each active hub receives either $\lfloor K\rfloor$ or $\lceil K\rceil$ slots through randomized rounding so that its expected capacity equals $K$.

\subsection{Scenario Initial States}

Because reachability depends on the optimism level's radius, the full nine-state initialization differs slightly across levels even though the underlying $N/E/M$ program counts are identical. The known ARC-supported pilot school is Early and reachable at all 12 levels. The table below gives the three anchor initializations; intermediate levels are calculated directly from their corresponding radii rather than by interpolating state counts.

\begin{table}[H]
\centering
\small
\begin{tabular}{@{}l r r r@{}}
\toprule
State & Level 1 & Level 6 & Level 12 \\
\midrule
$N_U$ & $1{,}761$ & $1{,}750$ & $1{,}732$ \\
$N_R$ & $29$ & $40$ & $58$ \\
$E_U$ & $43$ & $43$ & $43$ \\
$E_R$ & $0$ & $0$ & $0$ \\
$E_S$ & $1$ & $1$ & $1$ \\
$M_U$ & $82$ & $82$ & $79$ \\
$M_R$ & $9$ & $9$ & $12$ \\
$M_S$ & $0$ & $0$ & $0$ \\
$M_H$ & $0$ & $0$ & $0$ \\
\bottomrule
\end{tabular}
\end{table}

At every level, the school-state counts sum to $n=1{,}925$, with the same marginal program totals $N(0)=1{,}790$, $E(0)=44$, and $M(0)=91$. Only the initial reachable/unreachable partition changes with the assumed geographic radius. The college initialization remains $C_1(0)=1$ and $C_0(0)=31$ at all 12 levels.
\FloatBarrier
\section{Parameter Sensitivity Analysis}
\label{sec:parameter_sensitivity}

Several ARC-specific parameters cannot yet be estimated reliably from the trial deployment. We therefore tested how strongly projected outcomes depend on the parameter ranges represented by the 12 optimism levels. The main questions are which assumptions most strongly control projected growth and whether the same major drivers remain important across different combinations of assumptions.

We used two complementary sensitivity tests: one-at-a-time (OAT) sweeps and Morris screening \cite{morris1991factorial,campolongo2007effective}. OAT shows the direction and size of the effect produced by changing one factor around the level-6 scenario. Morris instead tests each factor under many different combinations of the other assumptions, showing which factors have the greatest influence more broadly. Using both lets us compare an easily interpreted scenario sweep with a global sensitivity test.

For OAT, we varied one sensitivity factor through optimism levels 1--12 while holding all other factors at level 6. We tested 11 factors: nine individual parameters and two parameter bundles. The bundles were Supported-program closure/backslide $(\delta_E^S,\nu_M^S,\delta_M^S)$ and secondary-hub closure/backslide $(\nu_M^H,\delta_M^H)$, because the rates within each bundle already vary together through a common multiplier in the scenario construction. Changing reach radius $R$ also recalculates the geographic relationships and initial reachable states derived from that radius. We measured the resulting change in projected outcomes at years 10, 20, and 40.

Morris repeatedly changes one factor by a fixed step and records how much the output changes, while starting from many different combinations of the other factors. We rank factors using $\mu^*$, the average absolute size of these changes, so larger values indicate greater overall influence. We also calculate $\sigma$, which becomes large when a factor's effect varies substantially across different surrounding assumptions. We used six grid levels and 30 Morris trajectories. Each parameter setting was averaged over 128 paired Monte Carlo realizations, using the same random-number streams within each comparison to reduce simulation noise.

The two tests produced closely aligned results. For projected schools with programs, both ranked $\kappa$, $\chi^B$, $\bar{w}^{\mathrm{new}}$, $\chi^S$, and the Supported closure/backslide bundle as the five strongest factors at year 10, in that order. At year 20, both gave the same leading six: $\chi^S$, $\kappa$, $\chi^B$, $\bar{w}^{\mathrm{new}}$, $K$, and $R$. At year 40, they retained the same leading six factors, with only $\chi^B$ and $\bar{w}^{\mathrm{new}}$ exchanging positions. Reach radius $R$ was the strongest factor for the number of schools within ARC reach at years 10, 20, and 40, while $\kappa$ was consistently strongest for primary-hub growth. Secondary-hub growth was driven most strongly by $\chi^B$ at year 10 and by $\chi^S$ at years 20 and 40. 

Together, these results show how ARC's growth drivers shift as the network develops. Early expansion is driven primarily by establishing additional primary hubs and converting existing Mature programs into secondary hubs, while later growth depends increasingly on Supported programs becoming further secondary hubs. The close agreement between OAT and Morris shows that these same major drivers remain important across many surrounding assumptions. Overall, ARC scales when it can successfully create and sustain new hubs, with direct support, geographic reach, and local capacity determining how far that hub-driven growth can propagate.

\begin{table}[t]
\centering
\small
\setlength{\tabcolsep}{5pt}
\caption{One-at-a-time sensitivity of projected schools with programs, $P(t)$. Each factor was varied from optimism level 1 to level 12 while all other factors remained fixed at level 6. Values report the change in projected programs from the level-1 to level-12 setting for that factor.}
\label{tab:program-sensitivity-oat}
\begin{tabular}{lrrr}
\toprule
Sensitivity factor & Year 10 & Year 20 & Year 40 \\
\midrule
Supported secondary-hub formation ($\chi^S$)
& 7.4 & 70.2 & 429.6 \\
Direct-support capacity per hub ($K$)
& 5.5 & 51.1 & 338.4 \\
Reach radius ($R$)
& 5.1 & 42.4 & 302.3 \\
New support-request probability ($p_{\mathrm{req}}^{\mathrm{new}}$)
& 9.1 & 52.4 & 216.4 \\
Unsupported secondary-hub formation ($\chi^B$)
& 14.4 & 57.9 & 213.3 \\
Primary-hub activation ($\kappa$)
& 19.6 & 63.7 & 165.9 \\
Supported closure/backslide bundle
($\delta_E^S,\nu_M^S,\delta_M^S$)
& 6.5 & 32.9 & 120.5 \\
Secondary-hub closure/backslide bundle
($\nu_M^H,\delta_M^H$)
& 1.7 & 17.6 & 109.7 \\
Secondary-hub retention ($\rho$)
& 1.4 & 12.9 & 93.8 \\
Supported Early-program maturation ($\mu_E^S$)
& 1.4 & 8.8 & 31.7 \\
Primary-hub cessation ($\xi$)
& 2.4 & 10.7 & 30.3 \\
\bottomrule
\end{tabular}
\end{table}

\begin{table}[t]
\centering
\small
\setlength{\tabcolsep}{5pt}
\caption{Morris sensitivity of projected schools with programs, $P(t)$. Values are $\mu^*$, the average absolute change produced by a Morris step on the common normalized optimism-level scale. Larger values indicate greater influence across the tested combinations of assumptions. Bold values mark the strongest factor at each year.}
\label{tab:program-sensitivity-morris}
\begin{tabular}{lrrr}
\toprule
Sensitivity factor & Year 10 & Year 20 & Year 40 \\
\midrule
Supported secondary-hub formation ($\chi^S$)
& 7.3 & \textbf{66.4} & \textbf{380.6} \\
Direct-support capacity per hub ($K$)
& 4.6 & 43.8 & 279.4 \\
Reach radius ($R$)
& 4.7 & 37.9 & 251.6 \\
New support-request probability ($p_{\mathrm{req}}^{\mathrm{new}}$)
& 8.1 & 45.4 & 174.3 \\
Unsupported secondary-hub formation ($\chi^B$)
& 12.8 & 49.0 & 183.7 \\
Primary-hub activation ($\kappa$)
& \textbf{16.8} & 53.3 & 138.6 \\
Supported closure/backslide bundle
($\delta_E^S,\nu_M^S,\delta_M^S$)
& 5.3 & 28.7 & 114.0 \\
Secondary-hub closure/backslide bundle
($\nu_M^H,\delta_M^H$)
& 1.8 & 15.8 & 91.9 \\
Secondary-hub retention ($\rho$)
& 1.0 & 10.4 & 74.6 \\
Supported Early-program maturation ($\mu_E^S$)
& 0.9 & 5.5 & 25.8 \\
Primary-hub cessation ($\xi$)
& 2.4 & 10.4 & 29.8 \\
\bottomrule
\end{tabular}
\end{table}
\FloatBarrier
\section{Extended Spatial Simulation Results}
\label{sec:extended_sim_results}

\begin{table}[H]
\centering
\small
\begin{tabular}{@{}lrrrr@{}}
\toprule
Scenario & Programs & Reachable & \shortstack{Primary\\hubs} & \shortstack{Secondary\\hubs} \\
\midrule
ARC off  & $161.4$     & $0.0$       & $0.0$ & $0.0$ \\
Level 1  & $338.0$     & $592.3$     & $5.7$ & $4.8$ \\
Level 2  & $393.1$     & $733.2$     & $6.2$ & $8.1$ \\
Level 3  & $477.1$     & $898.1$     & $6.7$ & $13.9$ \\
Level 4  & $601.0$     & $1{,}063.4$ & $7.2$ & $23.7$ \\
Level 5  & $772.8$     & $1{,}241.1$ & $7.6$ & $40.0$ \\
Level 6  & $992.2$     & $1{,}415.1$ & $8.0$ & $67.2$ \\
Level 7  & $1{,}134.0$ & $1{,}516.9$ & $8.4$ & $89.1$ \\
Level 8  & $1{,}271.7$ & $1{,}609.2$ & $8.6$ & $117.1$ \\
Level 9  & $1{,}393.8$ & $1{,}682.7$ & $8.9$ & $150.3$ \\
Level 10 & $1{,}491.9$ & $1{,}743.6$ & $9.2$ & $187.9$ \\
Level 11 & $1{,}572.8$ & $1{,}791.9$ & $9.6$ & $229.6$ \\
Level 12 & $1{,}638.0$ & $1{,}829.1$ & $9.9$ & $276.4$ \\
\bottomrule
\end{tabular}
\caption{Mean Indiana public-school outcomes after 40 years in the full spatial model. Each ARC optimism level is averaged over 1,000 Monte Carlo realizations.}
\label{tab:spatial_results_all}
\end{table}

\begin{figure}[H]
\centering
\includegraphics[width=\linewidth]{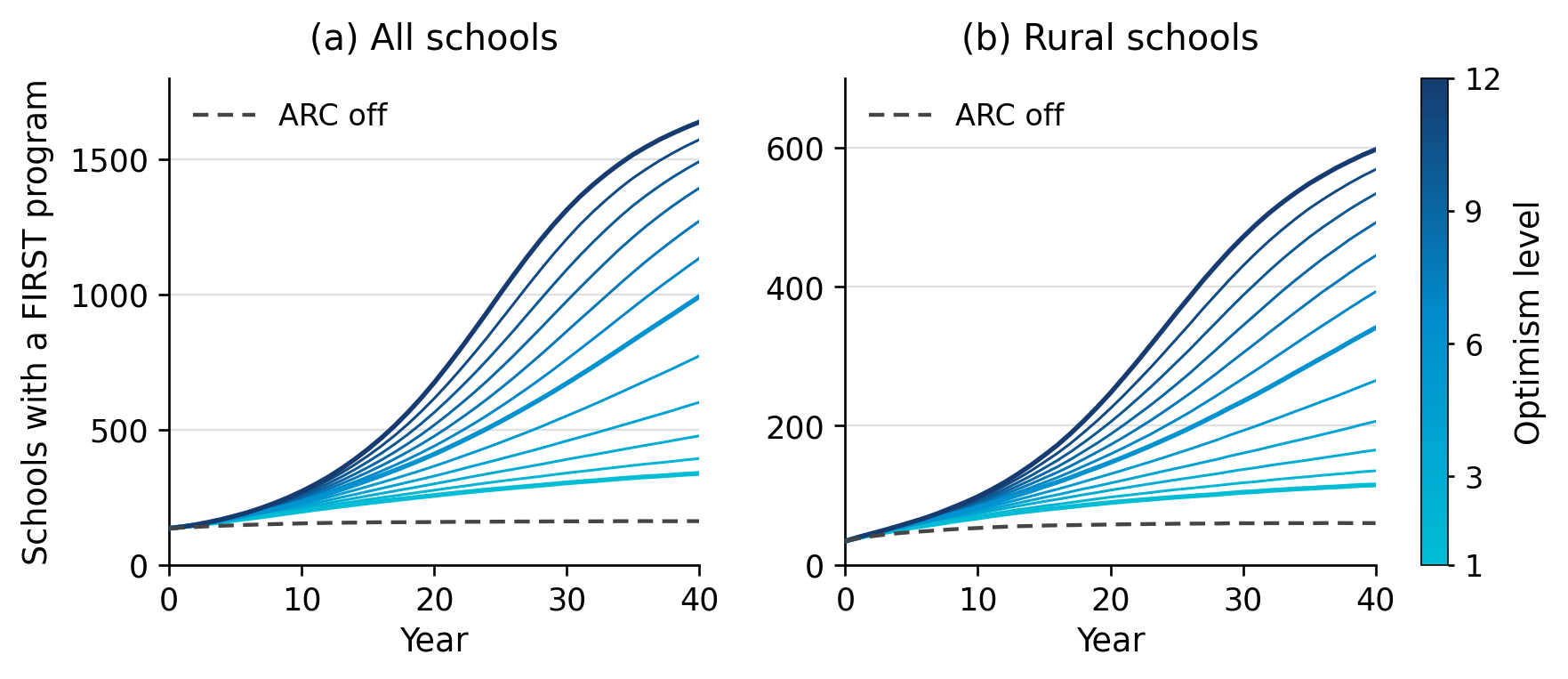}
\caption{Program growth for all 1,925 modeled schools (left) and the 719 rural schools (right) across all 12 optimism levels. Solid lines show Full ARC means over 1,000 Monte Carlo runs, with levels 1, 6, and 12 emphasized. Dashed lines show ARC off. Rural schools lie outside the 2020 Census urban-area boundaries used in Figure~\ref{fig:arc_promblem_and_loops}.}
\label{fig:rural_program_growth}
\end{figure}

\begin{table}[H]
\centering
\small
\setlength{\tabcolsep}{5pt}
\renewcommand{\arraystretch}{1.12}
\begin{tabular}{@{}clrrrrrr@{}}
\toprule
& & \multicolumn{2}{c}{Year 10} & \multicolumn{2}{c}{Year 20} & \multicolumn{2}{c}{Year 40} \\
\cmidrule(lr){3-4}\cmidrule(lr){5-6}\cmidrule(l){7-8}
Level & Model & Rural & Urban & Rural & Urban & Rural & Urban \\
\midrule
-- & ARC off & 52.9 & 100.2 & 58.0 & 100.1 & 59.9 & 101.5 \\
\midrule
1 & Full ARC & 67.4 & 130.8 & 89.6 & 167.0 & 115.0 & 223.0 \\
 & Primary hubs only & 64.6 & 124.4 & 80.6 & 147.6 & 93.1 & 174.0 \\
 & \shortstack[l]{Initial primary +\\secondary hubs} & 59.3 & 113.4 & 68.7 & 122.3 & 75.5 & 133.9 \\
\midrule
6 & Full ARC & 82.9 & 150.8 & 147.3 & 262.2 & 340.9 & 651.2 \\
 & Primary hubs only & 72.2 & 130.2 & 95.9 & 165.0 & 118.3 & 204.9 \\
 & \shortstack[l]{Initial primary +\\secondary hubs} & 68.5 & 124.0 & 99.0 & 170.5 & 189.9 & 360.6 \\
\midrule
12 & Full ARC & 98.1 & 173.7 & 247.3 & 426.6 & 597.4 & 1,040.5 \\
 & Primary hubs only & 78.1 & 137.7 & 109.9 & 184.7 & 143.1 & 241.2 \\
 & \shortstack[l]{Initial primary +\\secondary hubs} & 77.0 & 135.0 & 152.2 & 253.1 & 448.0 & 794.1 \\
\bottomrule
\end{tabular}
\caption{Mean numbers of rural and urban schools with FIRST programs after 10, 20, and 40 years, over 1,000 runs per scenario. All scenarios start with 34 rural and 101 urban programs. At level 6, Full ARC yields 341 rural programs after 40 years, compared with 118 with primary hubs only and 190 with the initial primary hub plus secondary hubs.}
\label{tab:rural_urban_programs}
\end{table}

\FloatBarrier
\section{Survey Information}
\label{sec:survey}

The evaluation used two anonymous end-of-program surveys: one for undergraduate mentors and one for parents or guardians of participating K--12 students. All respondents were at least 18 years old. Participation was voluntary, respondents could skip questions or stop at any time, and no names or other identifying information were collected. Surveys were administered via Qualtrics.

\subsection{Undergraduate Survey}

All nine students enrolled in the ARC course were eligible to participate, and seven completed the survey. The survey was distributed after the course ended and final grades had been submitted. Respondents rated both their perceived state before the course and their perceived state after the course during the same survey administration.

At the beginning of the survey, students were told the purpose of the study, that participation was voluntary and anonymous, and that participation would not affect their grades or academic standing. They were told that they could skip questions or stop at any time. Respondents confirmed that they were at least 18 years old and agreed to participate before beginning the survey.

\noindent\textbf{Question Types.} The survey contained 35 items after the consent screen. Eighteen items formed nine retrospective before-and-after pairs covering teaching, communication, group management, community connection, mentoring, leadership, and interest in AI, robotics, and graduate school. These items used a five-point Likert scale. The survey also included nine post-course rating questions, four background questions, three short-response questions, and one gold-standard attention check instructing respondents to select 4. All seven respondents answered the attention check correctly.

\noindent\textbf{Demographics.} The students came from several undergraduate years and included both computer science and engineering students. The class also included a mix of genders, countries of origin, and primary languages. Exact demographic information in these categories was not collected. Of the seven survey respondents, six selected ``CS/AI/DS Major'' and one selected ``Other.'' Two were first-year students, one was a second-year student, three were fourth-year students, and one did not report a year. Four reported prior FIRST experience and five reported prior teaching or mentoring experience.

\begin{table*}[ht]
\centering
\scriptsize
\begin{tabularx}{\textwidth}{@{}X p{0.24\textwidth}@{}}
\toprule
Survey question & Possible answers \\
\midrule

Before taking the course, I felt confident teaching technical concepts to students younger than myself. &
Likert (1--5) \\

After taking the course, I felt confident teaching technical concepts to students younger than myself. &
Likert (1--5) \\

Before taking the course, I felt able to adapt my explanations and communications when someone did not understand. &
Likert (1--5) \\

After taking the course, I felt able to adapt my explanations and communications when someone did not understand. &
Likert (1--5) \\

Before taking the course, I felt confident in managing and working in small groups. &
Likert (1--5) \\

After taking the course, I felt confident in managing and working in small groups. &
Likert (1--5) \\

Before taking the course, I felt connected to the local community. &
Likert (1--5) \\

After taking the course, I felt connected to the local community. &
Likert (1--5) \\

Before taking the course, I found teaching and mentoring to be enjoyable and meaningful. &
Likert (1--5) \\

After taking the course, I found teaching and mentoring to be enjoyable and meaningful. &
Likert (1--5) \\

Before taking the course, I was confident in my leadership skills. &
Likert (1--5) \\

After taking the course, I was confident in my leadership skills. &
Likert (1--5) \\

Select 4. &
Likert (1--5) \\

Before taking the course, I had a major interest in AI. &
Likert (1--5) \\

After taking the course, I had a major interest in AI. &
Likert (1--5) \\

Before taking the course, I had a major interest in Robotics. &
Likert (1--5) \\

After taking the course, I had a major interest in Robotics. &
Likert (1--5) \\

Before taking the course, I had a major interest in attending graduate school. &
Likert (1--5) \\

After taking the course, I had a major interest in attending graduate school. &
Likert (1--5) \\

Do you feel more connected with the community because of this course? &
Likert (1--5) \\

Do you feel you build good mentor-mentee relationships with the students? &
Likert (1--5) \\

Do you feel you have improved as a teacher/mentor due to this course? &
Likert (1--5) \\

Do you feel you have improved as a leader due to this course? &
Likert (1--5) \\

Do you feel this course made you more likely to pursue graduate study for research or teaching? &
Likert (1--5) \\

How did you feel about the guest lectures? &
Likert (1--5) \\

How did you feel about the weekly time with the FLL students? &
Likert (1--5) \\

How was the course instructor in teaching you mentoring skills. &
Likert (1--5) \\

How was the course instructor in teaching you about connecting AI, Robotics, and Community? &
Likert (1--5) \\

I am a ... &
[CS/AI/DS Major $|$ Other] \\

I am in year ... &
[1 $|$ 2 $|$ 3 $|$ 4] \\

I have prior experience in FIRST. &
[Yes $|$ No] \\

I have prior experience in teaching or mentoring. &
[Yes $|$ No] \\

Please write what you think this course did well. &
Short Response \\

Please write any suggestions you have to improve the course. &
Short Response \\

Please share any meaningful stories you have from your interactions with the FLL students. Please do not use real names or personal information. &
Short Response \\

\bottomrule
\end{tabularx}
\caption{Undergraduate survey questions and possible responses.}
\label{tab:undergrad_survey_questions}
\end{table*}

\subsection{Parent Proxy Survey}

K--12 students were not surveyed directly. Parents or legal guardians instead reported on their child's experience after the program. Four parents completed the survey. Parents could answer using their own observations, conversations with their child, or both.

At the beginning of the survey, parents were told that the study examined how participation in the AI and robotics outreach workshops may influence students' interest, confidence, and engagement in STEM, programming, and basic AI concepts. They were told that the survey was anonymous and voluntary, that they could skip questions or stop at any time, and that no data would be collected directly from children. Respondents confirmed that they were at least 18 years old and agreed to participate before beginning the survey.

\noindent\textbf{Question Types.} The survey contained 22 items after the consent screen. Twelve items formed six retrospective before-and-after pairs covering basic robot programming knowledge, access to programming resources, opportunities to practice programming and problem solving, enjoyment of robotics club, plans to continue robotics, and interest in further STEM education. These items used a five-point Likert scale from Strongly Disagree to Strongly Agree. Five additional Likert questions asked about the child's experience with the mentors and the effects of mentoring. The survey also contained two gold-standard attention checks, two short-response questions, and one question asking whether the parent had discussed the survey with their child. All four respondents answered both attention checks correctly. One parent reported discussing the survey with their child before answering. Two reported that they had not discussed it with their child. One reported that they had not discussed it with their child but had attended the competition near the end of the program.

\noindent\textbf{Demographics.} Detailed demographic information about the parents or children was not collected. All children were in 5th grade attending the rural elementary school focused on in the trial deployment, and were members of the school robotics club.

\begin{table*}[ht]
\centering
\scriptsize
\begin{tabularx}{\textwidth}{@{}X p{0.24\textwidth}@{}}
\toprule
Survey question & Possible answers \\
\midrule

Before the program, my child knew how to do basic robot programming. &
Likert (1--5) \\

After the program, my child knew how to do basic robot programming. &
Likert (1--5) \\

Before the program, my child's team had access to the programming resources they needed. &
Likert (1--5) \\

After the program, my child's team had access to the programming resources they needed. &
Likert (1--5) \\

Before the program, my child had opportunities to practice programming and problem-solving skills. &
Likert (1--5) \\

After the program, my child had opportunities to practice programming and problem-solving skills. &
Likert (1--5) \\

Before the program, my child enjoyed their robotics club. &
Likert (1--5) \\

After the program, my child enjoyed their robotics club. &
Likert (1--5) \\

For this question, please select ``Agree.'' &
Likert (1--5) \\

Before the program, my child planned to continue robotics club. &
Likert (1--5) \\

After the program, my child planned to continue robotics club. &
Likert (1--5) \\

Before the program, my child considered pursuing further education in science, robotics, or programming. &
Likert (1--5) \\

After the program, my child considered pursuing further education in science, robotics, or programming. &
Likert (1--5) \\

My child enjoyed working with the university mentors. &
Likert (1--5) \\

My child felt supported and encouraged by the university mentors. &
Likert (1--5) \\

The mentoring helped my child's FLL team prepare for competition. &
Likert (1--5) \\

The mentoring improved my child's ability to solve problems and try new ideas. &
Likert (1--5) \\

The mentoring improved teamwork on my child's FLL team. &
Likert (1--5) \\

For this question, please select ``Disagree.'' &
Likert (1--5) \\

How do you feel this program affected your child's interest in robotics and STEM going forward? &
Short Response \\

Did you discuss this survey with your child? &
[Yes, before answering $|$ No $|$ No, but I attended the competition] \\

Please share anything meaningful your child said about the mentors, the program, or the FLL season. Do not use names or personal details. &
Short Response \\

\bottomrule
\end{tabularx}
\caption{Parent-proxy survey questions and possible responses.}
\label{tab:parent_survey_questions}
\end{table*}
\FloatBarrier
\section{Survey Response Data}
\label{sec:survey_open}

Because of the small number of respondents, Tables~\ref{tab:undergrad_raw}
and~\ref{tab:parent_raw} report the complete individual Likert responses
underlying the retrospective pre--post results. Responses are listed in a
fixed respondent order so that before and after values remain paired.
All items use a 1--5 scale from Strongly Disagree to Strongly Agree.

\begin{table}[ht]
\centering
\small
\caption{Complete undergraduate retrospective pre--post Likert responses
($n=7$). Entries within each cell correspond to respondents U1--U7 in the
same order. An em dash indicates a skipped response.}
\label{tab:undergrad_raw}
\begin{tabular}{lcc}
\toprule
Measure & Before & After \\
\midrule
Teaching technical concepts
    & 2, 4, 4, 4, 3, 1, 3
    & 4, 4, 5, 4, 5, 4, 4 \\
Adapting explanations
    & 3, 3, 4, 4, 2, 2, 5
    & 4, 4, 5, 4, 4, 5, 5 \\
Small-group management
    & 4, 2, 4, 3, 3, 1, 4
    & 4, 3, 5, 4, 4, 4, 4 \\
Community connection
    & 1, 1, 4, 2, 1, 2, 2
    & 3, 3, 5, 4, 5, 4, 4 \\
Teaching/mentoring meaning
    & 2, 3, 4, 4, 4, 4, 3
    & 4, 5, 5, 4, 5, 5, 4 \\
Leadership confidence
    & 3, 1, 4, --, 3, 2, 4
    & 4, 1, 5, 3, 4, 4, 4 \\
AI interest
    & 1, 3, 5, 3, 5, 3, 5
    & 2, 3, 5, 4, 5, 4, 5 \\
Robotics interest
    & 5, 4, 5, 4, 5, 2, 4
    & 5, 4, 5, 4, 5, 4, 5 \\
Graduate-school interest
    & 4, 1, 5, 3, 4, 3, 5
    & 4, 3, 5, 3, 5, 4, 5 \\
\bottomrule
\end{tabular}
\end{table}

\begin{table}[ht]
\centering
\small
\caption{Complete parent-proxy retrospective pre--post Likert responses
($n=4$). Entries within each cell correspond to respondents P1--P4 in the
same order.}
\label{tab:parent_raw}
\begin{tabular}{lcc}
\toprule
Measure & Before & After \\
\midrule
Robot programming knowledge
    & 1, 4, 2, 1
    & 4, 5, 2, 5 \\
Access to programming resources
    & 2, 3, 2, 1
    & 4, 5, 4, 4 \\
Programming/problem-solving practice
    & 3, 4, 2, 3
    & 4, 5, 3, 5 \\
Enjoyed robotics club
    & 4, 3, 3, 3
    & 5, 5, 2, 5 \\
Planned to continue robotics club
    & 4, 5, 3, 3
    & 4, 5, 2, 5 \\
Further STEM/programming education
    & 3, 5, 3, 2
    & 3, 5, 2, 4 \\
\bottomrule
\end{tabular}
\end{table}

\begin{longtable}{@{}p{0.29\textwidth}p{0.67\textwidth}@{}}
\caption{Undergraduate mentor open-response survey data.}
\label{tab:undergrad_open_responses}\\
\toprule
\textbf{Survey question} & \textbf{Response} \\
\midrule
\endfirsthead

\toprule
\textbf{Survey question} & \textbf{Response} \\
\midrule
\endhead

\bottomrule
\multicolumn{2}{@{}p{0.96\textwidth}@{}}{\footnotesize
\textit{Note.} Black bars indicate text removed for anonymity. An em dash indicates that no response was provided.} \\
\endlastfoot

Please write what you think this course did well. &
Introducing me to more applications of AI and robotics within the community, it gave me a lot more experience mentoring than I previously had. \\
\addlinespace

&
Multiple mentors per group worked well because it was hard to have one keep all the kids together in one area and not have at least one of them roaming. \\
\addlinespace

&
The facilitation was fantastic. The fact that the kids actually came here, we had two tables and several robots to work with, and everyone was able to go through the process smoothly is already a success. I also think the integration of mentoring and guest lectures from faculty at the forefront of their respective fields exposes students to a diverse range of new perspectives. \\
\addlinespace

&
I think this course connected to the community well. \\
\addlinespace

&
Definitely made me feel more connected to the local community \\
\addlinespace

&
I thought this course did a great job helping students prepare for the FLL competition because it supported growth in both the participants and the mentors. The weekly structure worked well, especially with hands-on sessions followed by guest lectures. Those talks helped me understand how AI and robotics appear in everyday life, and I realized that building LEGO robots and teaching them simple commands is the first step toward how these systems function in the real world. \\
\addlinespace

&
The guest lecturers sharing their research was really informative. \\
\midrule

Please write any suggestions you have to improve the course. &
Some of the planning for activities and group structures seems like it should be a little more concrete, which probably results from the fact it's a new course. \\
\addlinespace

&
More structure for the mentoring to make better use of the limited time with the students and strategies to be able to make the students want to do something instead of forcing them to do something like strategy before jumping in. It would also be nice to have a bigger space because the room we are in got really crowded and of course would be nice if we could have the table set up permanently. \\
\addlinespace

&
I think it should have been clearer from the beginning that we were preparing the students for a serious competition. We at \rule{1.2cm}{1.1ex} as well as the teachers at \rule{1.2cm}{1.1ex} learned a lot from this first year and I expect the team will do much better next year, especially considering how late in the season we started. \\
\addlinespace

&
I would say that providing more time with the children on the robotics teams to work on the robots more would be beneficial. In addition, having them focus on developing a plan and not developing randomly in every direction would also help. Mock competitions would also be highly beneficial. \\
\addlinespace

&
--- \\
\addlinespace

&
It could be helpful to have clearer pacing or checkpoints throughout the semester. Sometimes activities felt rushed or uneven, so having weekly goals or mini milestones when working with the kids and their robots would make it easier for both students and mentors to stay on track. We did this towards the end, but starting off with that could possibly work. \\
\addlinespace

&
The sessions with the students could have been geared more towards evaluation and feedback instead of building and programming. \\
\midrule

Please share any meaningful stories you have from your interactions with the FLL students. Please do not use real names or personal information. &
I think it was the third session with them, and I was supervising as the team rebuilt their robot, and while they were building, one of them began to ask me about different concepts regarding gear ratios and getting a certain part to move faster. I explained it as simply as I could, and I believe he gained a bit more of an understanding regarding that, which I found very meaningful. Also, by the end of the competition, we managed to get one of the children who had been trying to do more than he should to take a step back to help his other team members learn as he guided them through different parts. \\
\addlinespace

&
--- \\
\addlinespace

&
One student was clearly very invested in the competition, and also really wanted to do things his way. On the day of the competition, when things weren't working quite as he planned, I encouraged him to try something different, and collaborate with one of his teammates who wanted to help. Regardless of their final performance, seeing them work together for the good of the team showed clear growth. \\
\addlinespace

&
The kids told me after their last time in the room at \rule{1.2cm}{1.1ex} that they didn't think they would be there without me or my help. It really made me feel connected, and more ``a part of their team'' than I had previously. \\
\addlinespace

&
After the competition. One of the kid I’ve been working with told me that he will see me next year, it showed me that he not only learned the technical skills --- he also felt supported and wanted to keep going. That small moment made the whole experience worthwhile. \\
\addlinespace

&
I really enjoyed watching how their thinking process worked. If something did not go as planned, they would detect their mistake, try again, and sometimes even draw out where they wanted the robot to go to fix the issue. Seeing them do that made me realize strategies I had not considered before. \\
\addlinespace

&
--- \\

\end{longtable}

\begin{longtable}{@{}p{0.29\textwidth}p{0.67\textwidth}@{}}
\caption{Parent open-response survey data.}
\label{tab:parent_open_responses}\\
\toprule
\textbf{Survey question} & \textbf{Response} \\
\midrule
\endfirsthead

\toprule
\textbf{Survey question} & \textbf{Response} \\
\midrule
\endhead

\bottomrule
\multicolumn{2}{@{}p{0.96\textwidth}@{}}{\footnotesize
\textit{Note.} Black bars indicate text removed for anonymity. An em dash indicates that no response was provided.} \\
\endlastfoot

Please share anything meaningful your child said about the mentors, the program, or the FLL season. Do not use names or personal details. &
--- \\
\addlinespace

&
My son was impressed with his mentors and enjoyed working with them. \\
\addlinespace

&
He loved being with the mentors and his time at \rule{1.2cm}{1.1ex}. \\
\addlinespace

&
--- \\
\midrule

How do you feel this program affected your child's interest in robotics and STEM going forward? &
Positively \\
\addlinespace

&
My son has always had a strong interest in STEM and robotics and really enjoyed working with his \rule{1.2cm}{1.1ex} mentors and was sad the program had to end. I think it was a great experience for the kids to be able to work with college students and learn from their experience and just have the opportunity to communicate with a young adult. \\
\addlinespace

&
Unfortunately my child was very interested in programming but did not get the chance. That had nothing to do with the mentors. \\
\addlinespace

&
--- \\

\end{longtable}
\FloatBarrier
\section{ARC Course Details and Reproducibility}
\label{sec:course_reproducibility}

The ARC course was a three-credit undergraduate course meeting once per week for a two-hour block. Mentoring was the central course activity. Mentoring and participation accounted for 70\% of the course grade and required consistent attendance, communication with team coaches, and contribution to the teams' progress. A final essay accounted for 20\% and a final presentation for 10\%. Optional biweekly reflections asked students to connect course topics to their experiences and communities. These were eligible for extra credit. The final assignment required a 4--8 page essay and an 8--10 minute presentation connecting course topics with the student's mentoring experience. Students could write either a persuasive essay on AI, robotics, and community or a short research proposal addressing a community-relevant problem.

The first four weeks prepared the undergraduates before they began mentoring K--12 teams. Students were introduced to the FLL platform by building and programming small robots, learned about robotics education and FIRST, completed required youth-protection training, and practiced methods for teaching programming. The first K--12 workshop occurred in week five. After workshops began, approximately 90 minutes of each weekly meeting were devoted to direct mentoring and approximately 30 minutes to continuing course material. Table~\ref{tab:arc_course_schedule} summarizes the course structure.

\begin{table*}[t]
\centering
\small
\begin{tabular}{@{}clp{0.72\textwidth}@{}}
\toprule
Week & Phase & Topic / activity \\
\midrule
1 & Preparation & Course introduction; FLL kits; build a small mobile robot. \\
2 & Preparation & Robotics education and FIRST; youth-protection training; FLL programming. \\
3 & Preparation & Practical methods for teaching programming; FLL programming. \\
4 & Preparation & Practical teaching exercise. \\
5 & Workshops begin & Workshop preparation and first K--12 mentoring session. \\
6--10 & Workshops + seminars & Recurring mentoring; seminars on citizen science, robotics safety and ethics, FTC/FRC, human--robot teaming, and AI literacy. \\
11--14 & Workshops + seminars & Continued mentoring; later topics included autonomous systems, AI safety and explainability, and preparation for competition and final work. \\
15--16 & Synthesis & Final-project support and student presentations. \\
\bottomrule
\end{tabular}
\caption{Structure of the ARC course during the initial deployment. Seminar topics can be replaced with locally relevant AI and robotics topics while retaining the mentor-training and workshop structure.}
\label{tab:arc_course_schedule}
\end{table*}

A primary hub does not need to reproduce the trial course lecture-for-lecture. The essential components are mentor preparation followed by recurring, supervised K--12 workshops. At least one instructor or other experienced robotics mentor is needed to prepare the undergraduate mentors and remain available during workshops. This role cannot be replaced entirely by fixed lesson material because teams differ in technical problems, pace, competition readiness, and group dynamics. The experienced mentor must be able to help when a team or novice mentor encounters a problem that was not anticipated in the course material.

The physical requirements depend strongly on the robotics program being supported. For the FLL deployment studied here, a hub needs FLL robot kits, devices capable of programming them, the current competition field and game materials, sufficient working space, supervision, and the required youth-protection procedures. The trial provided two FLL game arenas so that multiple teams could test robots during the same workshop. The same course structure can be used with FTC or FRC teams, but those programs require progressively greater hardware, workspace, fabrication and maintenance resources, and transportation support. For FTC and FRC, basic safety training should also be mandatory for all mentors, as these programs commonly use power tools.

The continuing seminar component is intentionally portable. The trial used both instructor-led and guest seminars to connect robotics mentoring with AI and robotics research and with problems affecting the surrounding community. Another university does not need the same lecturers or presentation materials. Local faculty, researchers, robotics organizations, and community partners can instead provide topics relevant to that region. These topics should focus on real research or development in AI or robotics, and how that development can impact local communities.

\end{document}